\documentclass{article}
 \usepackage[nonatbib,preprint]{neurips_2022}

\usepackage[utf8]{inputenc} % allow utf-8 input
\usepackage[T1]{fontenc}    % use 8-bit T1 fonts
\usepackage{hyperref}       % hyperlinks
\hypersetup{colorlinks,urlcolor=blue}
\usepackage{url}            % simple URL typesetting
\usepackage{booktabs}       % professional-quality tables
\usepackage{amsfonts}       % blackboard math symbols
\usepackage{nicefrac}       % compact symbols for 1/2, etc.
\usepackage{microtype}      % microtypography
\usepackage{xcolor}         % colors
\usepackage{caption}
\usepackage[ruled,vlined,linesnumbered]{algorithm2e}
\usepackage{amsthm}
\usepackage{graphicx}
\usepackage{amsmath,amssymb}
\usepackage{array}
\usepackage{longtable}
\usepackage{subfigure}
\usepackage{cleveref}
\usepackage{graphicx}
\usepackage{todonotes}
\usepackage{wrapfig}
\usepackage{needspace}
\usepackage{booktabs}
\usepackage{changepage} % to change margins in large tables
\usepackage{xspace}

\newcommand{\BYOL}{\texttt{BYOL}\xspace}
\newcommand{\BYOLE}{\texttt{BYOL-Explore}\xspace}
\newcommand{\ICM}{\texttt{ICM}\xspace}
\newcommand{\RDD}{\texttt{R2D2}\xspace}
\newcommand{\nethack}{\texttt{NetHack}\xspace}
\newcommand{\DMH}{\texttt{DM-HARD-8}\xspace}

\newcommand{\RND}{\texttt{RND}\xspace}
\newcommand{\CommaBin}{\mathbin{\raisebox{0.5ex}{,}}}

\renewcommand{\epsilon}{\varepsilon}

\let\originalleft\left
\let\originalright\right
\renewcommand{\left}{\mathopen{}\mathclose\bgroup\originalleft}
\renewcommand{\right}{\aftergroup\egroup\originalright}

\title{\BYOLE:\\ Exploration by Bootstrapped Prediction}

\author{%
  Zhaohan Daniel Guo\thanks{Equal contribution.}\\
  DeepMind \\
  danielguo@deepmind.com \\
  \And
  Shantanu Thakoor$^*$ \\
  DeepMind \\
  \And
  Miruna P\^{i}slar$^*$ \\
  DeepMind \\
  \And
  Bernardo Avila Pires$^*$ \\
  DeepMind \\
  \And
  Florent Altch\'e$^*$ \\
  DeepMind \\
  \And
  Corentin Tallec$^*$ \\
  DeepMind \\
  \And
  Alaa Saade \\
  DeepMind \\
  \And
  Daniele Calandriello \\
  DeepMind \\
  \And
  Jean-Bastien Grill \\
  DeepMind \\
  \And
  Yunhao Tang \\
  DeepMind \\
  \And
  Michal Valko \\
  DeepMind \\
  \And
  R\'emi Munos \\
  DeepMind \\
  \And
  Mohammad Gheshlaghi Azar$^*$ \\
  DeepMind \\
  mazar@deepmind.com\\
  \And
  Bilal Piot$^*$ \\
  DeepMind \\
  piot@deepmind.com
}

\begin{document}

\maketitle

\begin{abstract}

We present \BYOLE, a conceptually simple yet general approach for curiosity-driven exploration in visually-complex environments. \BYOLE learns a world representation, the world dynamics, and an exploration policy all-together by optimizing a single prediction loss in the latent space with no additional auxiliary objective. We show that \BYOLE is effective in \DMH, a challenging partially-observable continuous-action hard-exploration benchmark with visually-rich $3$-D environments. On this benchmark, we solve the majority of the tasks purely through augmenting the extrinsic reward with \BYOLE's intrinsic reward, whereas prior work could only get off the ground with human demonstrations. As further evidence of the generality of \BYOLE, we show that it achieves superhuman performance on the ten hardest exploration games in \texttt{Atari} while having a much simpler design than other competitive agents.
\end{abstract}

\section{Introduction}
\label{sec:introduction}
Exploration is essential to \textit{reinforcement learning} (RL)~\cite{sutton1998reinforcement}, especially when extrinsic rewards are sparse or hard to reach. In rich environments, the variety of meaningful directions of exploration makes it impractical to visit everything. Thus, the question becomes: how can an agent determine which parts of the environment are interesting to explore? One promising paradigm to address this challenge is curiosity-driven exploration. It consists of (i) learning a predictive model of some information about the world, called a \textit{world model}, and (ii) using discrepancies between predictions of the world model and real experience to build intrinsic rewards~\cite{schmidhuber1991curious, sun2011planning, schmidhuber2010formal, houthooft2016variational, pathak2017curiosity, pathak2019self, azar2019world}. An RL agent optimizing these intrinsic rewards drives itself towards states where the world model is incorrect or imperfect, generating new trajectories on which the world model can be improved. In other words, the properties of the world model influence the quality of the exploration policy, which in turn gathers new data to shape the world model itself. Thus, it can be important not to treat learning the world model and learning the exploratory policy as two separate problems, but instead altogether as a single joint problem to solve. 

In this paper, we present \BYOLE, a curiosity-driven exploration algorithm whose appeal resides in its conceptual simplicity, generality, and high performance. \BYOLE learns a world model with a self-supervised prediction loss, and uses the same loss to train a curiosity-driven policy, thus using a single learning objective to solve both the problem of building the world model's representation and the curiosity-driven policy. Our approach builds upon \textit{Bootstrap Your Own Latent} (\BYOL), a latent-predictive self-supervised method which predicts an older copy of its own latent representation. This bootstrapping mechanism has already been successfully applied in computer vision~\cite{grill2020bootstrap, richemond2020byol}, graph representation learning~\cite{thakoor2022large}, and representation learning in RL~\cite{guo2020bootstrap,schwarzer2020data}. However, the latter works focus primarily on using the world-model for representation learning in RL whereas \BYOLE takes this one step further, and not only learns a versatile world model but also uses the world model's loss to drive exploration.

We evaluate \BYOLE on \DMH~\cite{gulcehre2019making}, a suite of $8$ complex first-person-view $3$-D tasks with sparse rewards. These tasks demand efficient exploration since in order to reach the final goal and obtain the reward  they require  completing a sequence of precise, orderly interactions with the physical objects in the environment, unlikely to happen under a  vanilla random exploration strategy (see Fig.~\ref{fig:dmh_human_knock_obj_demo} and~\href{https://drive.google.com/drive/folders/1ERMPmP_UIqCNNhGgHO7Pu3IwX0ESQdn2?usp=sharing}{videos}). To show the generality of our method we also evaluate \BYOLE on the ten hardest exploration \texttt{Atari} games~\cite{bellemare2016unifying}. In all these domains, \BYOLE outperforms other prominent curiosity-driven exploration methods, such as \textit{Random Network Distillation} (\RND)~\cite{burda2019exploration} and \textit{Intrinsic Curiosity Module} (\ICM)~\cite{pathak2017curiosity}. In \DMH, \BYOLE achieves human-level performance in the majority of the tasks using only the extrinsic reward augmented with \BYOLE's intrinsic reward, whereas previously significant progress required human demonstrations~\cite{gulcehre2019making}. Remarkably, \BYOLE achieves this performance using only a single world model and a single policy network concurrently trained across all tasks. Finally, as further evidence of its generality, \BYOLE achieves superhuman performance in the ten hardest exploration \texttt{Atari} games~\cite{bellemare2016unifying} while having a simpler design than other competitive agents, such as \texttt{Agent57}~\cite{badia2020agent57, badia2020never} and \texttt{Go-Explore}~\cite{ecoffet2019go, ecoffet2020first}.\footnote{Contrary to \texttt{Agent57}, \BYOLE  neither requires episodic memory nor using an additional bandit mechanism to mix long-term and short-term rewards. As opposed to \texttt{Go-Explore}, we do not have to explicitly keep in memory a set of diverse goal-states to visit, which requires setting additional hyper-parameters that are environment-dependent.}  

\section{Related Work}
\label{sec:relatedwork}

There is a large body of research in building world models either for planning~\cite{sun2011planning, sekar2020planning, hafner2019learning, hafner2019dream, schrittwieser2020mastering}, representation learning~\cite{schwarzer2020data, guo2020bootstrap, laskin2020curl, gregor2019shaping} or curiosity-driven exploration~\cite{schmidhuber1991curious, tang2017exploration, schmidhuber2010formal, houthooft2016variational, pathak2017curiosity, pathak2019self, azar2019world, sekar2020planning}. Most works consider world models that predict the entire observations~\cite{schmidhuber1991learning, oh2015action, finn2016unsupervised, gregor2019shaping}, which necessitates a loss in pixel space when observations are visually complex images. Some works have considered predicting latent representations, whether they are random projections~\cite{burda2018large,burda2019exploration}, or learned representations from a separate model, such as an inverse dynamics model~\cite{pathak2017curiosity} or an auto-encoder~\cite{ha2018world, burda2018large}. Finally, some RL works~\cite{schrittwieser2020mastering} have focused on predicting lower-dimensional quantities such as the extrinsic reward, the action-selection policy, and the value function to build a world model. 

Our \BYOLE's world model operates in latent space and uses the same loss both for representation and intrinsic reward, simplifying and unifying representation learning and exploration. \BYOLE's world model is derived from recent self-supervised representation learning methods~\cite{grill2020bootstrap, richemond2020byol, recasens2021broaden, thakoor2022large} and is similar to the ones in self-supervised RL~\cite{schwarzer2020data, guo2020bootstrap}. These previous works focused on the benefit of shaping representations for policy learning and have not looked into exploration. We build on this previous work to show that we can take the impact of a good representation technique further and use it to drive exploration.

While our approach belongs to the curiosity-driven exploration paradigm~\cite{oudeyer2007intrinsic,lopes2012exploration, oudeyer2009intrinsic, schmidhuber1991curious, bellemare2016unifying, tang2017exploration, schmidhuber2010formal, houthooft2016variational, pathak2017curiosity, pathak2019self, azar2019world, sekar2020planning}, other exploration paradigms have also been proposed. The maximum entropy  paradigms try to steer the agent to a desired distribution of states (or state-action pairs) that maximizes the entropy of visited states~\cite{hazan2019provably, tarbouriech2019active, tarbouriech2020active, guo2021geometric}. The goal-conditioned paradigm has the agent set its own goal drive exploration~\cite{schaul2015universal, andrychowicz2017hindsight, florensa2018automatic, warde2018unsupervised, nair2018visual, colas2019curious, zhao2019maximum, hartikainen2019dynamical, ecoffet2020first, pong2020skew,zhang2020automatic,pitis2020maximum}. The reward-free exploration paradigm consists of training an agent to explore the environment such that it would be able to produce a near-optimal policy for \textit{any} possible reward function~\cite{jin2020reward, kaufmann2021adaptive, menard2021fast,zanette2020provably, wang2020reward, chen2021near,zhang2021reward,zhang2020nearly}.

\section{Method}
\label{sec:method}
Our agent has three components: a self-supervised latent-predictive world-model called \BYOLE, a generic reward normalization and prioritization scheme, and an off-the-shelf RL agent that can optionally share its own representation with \BYOLE's world model.

\subsection{Background and Notation}
\label{sec:background and notations}
We consider a discrete-time interaction process~\cite{mccallum1995instance, hutter2004universal, hutter2009feature, daswani2013q} between an agent and its environment where, at each time step $t\in\mathbb{N}$, the agent receives an observation $o_t\in\mathcal{O}$ and generates an action $a_t\in\mathcal{A}$. We consider an environment with stochastic dynamics $p: \mathcal{H}\times\mathcal{A}\rightarrow\Delta_\mathcal{O}$\footnote{We write $\Delta_\mathcal{Y}$ the set of probability distributions over a set $\mathcal{Y}$.} that maps 
a history of past observations-actions and a current action to a probability distribution over future observations. More precisely, the space of past observations-actions is $\mathcal{H}=\bigcup_{t\in\mathbb{N}}\mathcal{H}_t$ where $\mathcal{H}_0=\mathcal{O}$ and $\forall t\in\mathbb{N}^*, \mathcal{H}_{t+1}=\mathcal{H}_t\times\mathcal{A}\times\mathcal{O}$. We consider policies $\pi: \mathcal{H}\rightarrow \Delta_\mathcal{A}$ that maps a history of past observations-actions to a probability distribution over actions. Finally, an extrinsic reward function $r_e: \mathcal{H}\times\mathcal{A}\rightarrow \mathbb{R}$ maps a history of past observations-actions to a real number.  

\subsection{Latent-Predictive World Model}
\BYOLE world model is a multi-step predictive world model operating at the latent level. It is inspired by the self-supervised learning method \BYOL in computer vision and adapted to interactive environments (see Section~\ref{sec:background and notations}). 
Similar to \BYOL, \BYOLE model trains an online network using targets generated by an exponential moving average (EMA) target network. However, \BYOL obtains its targets by applying different augmentations to the same observation as the online representation, whereas \BYOLE model gets its targets from future observations processed by an EMA of the online network, with no hand-crafted augmentation. Also \BYOLE model, uses a recurrent neural network (RNN)~\cite{hochreiter1997long, chung2014empirical} to build the agent state, i.e., the state of RNN,  from the history of observations, whereas the original \BYOL only uses a feed-forward network for encoding the observations. In the remainder of this section, we will explain: (i) how the online network builds future predictions, (ii) how targets for our predictions are obtained through a target network, (iii) the loss used to train the online network, and (iv) how we compute the uncertainties of the world model.

\begin{figure}[htbp]
    \centering
    \subfigure{\includegraphics[width=0.85\textwidth]{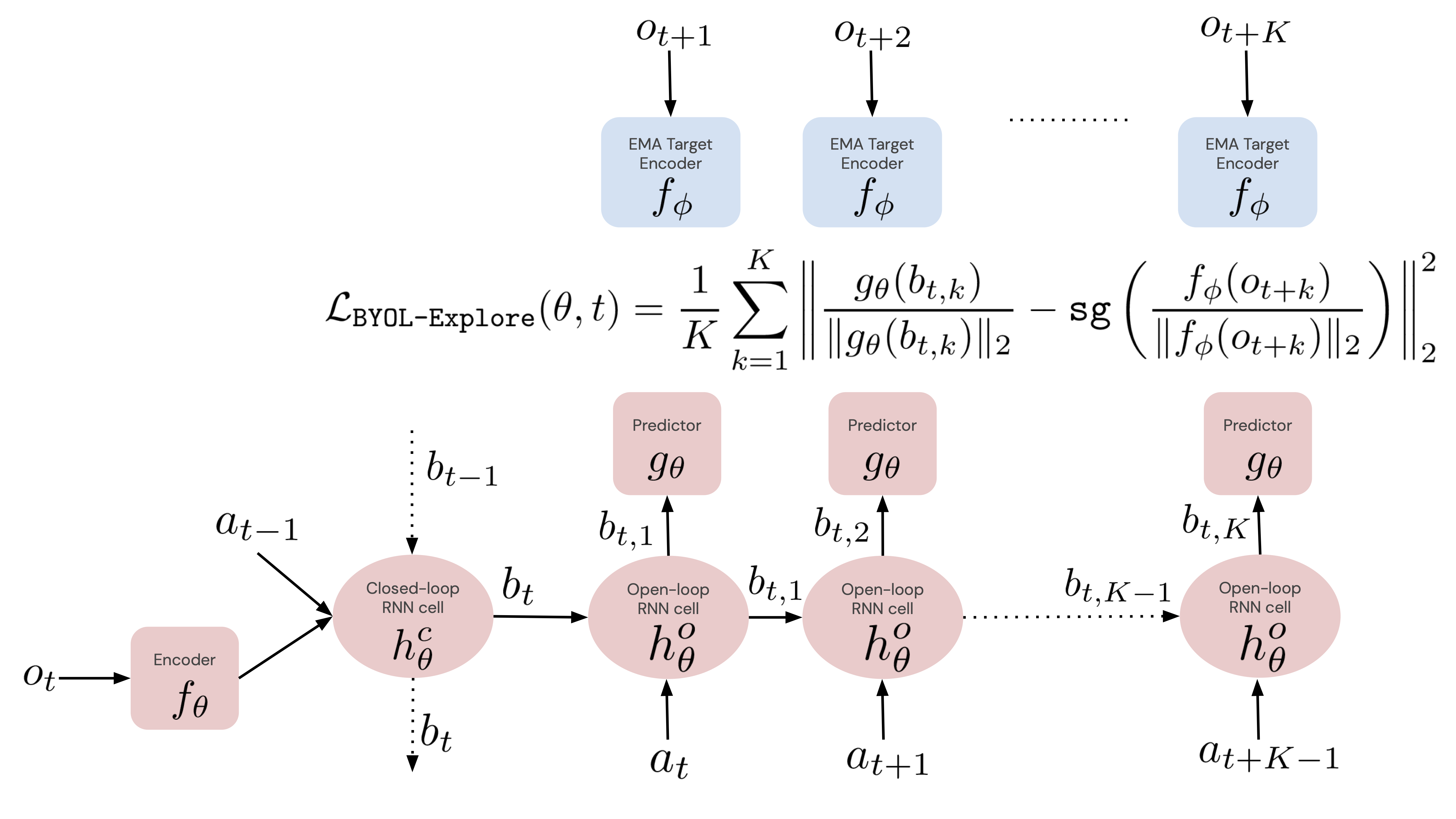}}
    \caption{\BYOLE's Neural Architecture (see main text for details).}
    \label{fig:byol_architecture}
\end{figure}

\paragraph{(i) Future Predictions.}
The online network is composed of an encoder $f_\theta$ that transforms an observation $o_t$ into an observation-representation $f_\theta(o_t)\in\mathbb{R}^N$, where $N\in\mathbb{N^*}$ is the embedding size. The observation-representation $f_\theta(o_t)$ is then fed alongside the previous action $a_{t-1}$ to a RNN cell $h^c_\theta$ that is referred as the close-loop RNN cell. It computes a representation $b_t\in\mathbb{R}^M$ of the history $h_t\in\mathcal{H}_t$ seen so far as $b_t = h^c_\theta(b_{t-1},a_{t-1}, f_\theta(o_t))$, where $M\in\mathbb{N}^*$ is the size of the history-representation. Then, the history-representation $b_t$ is used to initialize an open-loop RNN cell $h^o_\theta$ that outputs open-loop representations $(b_{t,k}\in\mathbb{R}^M)_{k=1}^{K-1}$ as $b_{t,k}=h^o_\theta(b_{t, k-1}, a_{t+k-1})$ where $b_{t, 0}=b_t$ and $K$ is the open-loop horizon. The role of the open-loop RNN cell is to \textit{simulate} future history-representations while observing only the future actions. Finally, the open-loop representation $b_{t,k}$ is fed to a predictor $g_\theta$ to output the open-loop prediction $g_\theta(b_{t,k})\in\mathbb{R}^N$ at time $t+k$ that plays the role of our future prediction at time $t+k$.

\paragraph{(ii) Targets and Target Network.}
The target network is an observation encoder $f_\phi$ whose parameters are an EMA of the online network's parameters $\theta$. It outputs targets $f_\phi(o_{t+k})\in\mathbb{R}^N$ that are used to train the online network. After each training step, the target network's weights are updated via an EMA update $\phi\leftarrow \alpha\phi + (1-\alpha)\theta$ where $\alpha$ is the target network EMA parameter. A sketch of the neural architecture is provided in Fig.~\ref{fig:byol_architecture}, with more details in App.~\ref{app:byol_archi}. 

\paragraph{(iii) Online Network Loss Function.} Suppose our RL agent collected a batch of trajectories $\left((o^j_t, a^j_t)_{t=0}^{T-1}\right)_{j=0}^{B-1}$, where $T\in\mathbb{N}^*$ is the trajectory length and $B\in\mathbb{N}^*$ is the batch size. Then, the loss $\mathcal{L}_{\BYOLE}(\theta)$ to minimize is defined as the average cosine distance between the open-loop future predictions $g_\theta(b^j_{t, k})$ and their respective targets $f_\phi(o^j_{t+k})$ at time $t+k$:
\begin{align*}
\mathcal{L}_{\BYOLE}(\theta, j, t, k)&= \left\|\frac{g_\theta(b^j_{t, k})}{\|g_\theta(b^j_{t, k})\|_2}-\texttt{sg}\left(\frac{f_\phi(o^j_{t+k})}{\|f_\phi(o^j_{t+k})\|_2}\right)\right\|_2^2,\\
\mathcal{L}_{\BYOLE}(\theta) &= \frac{1}{B(T-1)}\sum_{j=0}^{B-1}\sum_{t=0}^{T-2}\frac{1}{K(t)}\sum_{k=1}^{K(t)}\mathcal{L}_{\BYOLE}(\theta,j,t,k),    
\end{align*}
where $K(t)=\min(K, T-1-t)$ is the valid open-loop horizon for a trajectory of length $T$ and \texttt{sg} is the stop-gradient operator.

\paragraph{(iv) World Model Uncertainties}
The uncertainty associated to the transition $(o^j_t, a^j_t, o^j_{t+1})$ is the sum of the corresponding prediction losses:
\begin{equation*}
\ell^{\,j}_t=\sum_{p+q=t+1}\mathcal{L}_{\BYOLE}(\theta,j, p, q),    
\end{equation*}
where $0\leq p\leq T-2$,  $1\leq q\leq K$ and $0\leq t\leq T-2$. This accumulates all the losses corresponding to the world-model uncertainties relative to the observation $o^j_{t+1}$. Thus, a timestep receives intrinsic reward based on how difficult its observation was to predict from past partial histories.

\paragraph{Intuition on why \BYOLE  learns a meaningful representation.}
The intuition behind \BYOLE is similar in spirit to the one behind \BYOL.
In early training, the target network is initialized randomly, and so \BYOLE's online network and the closed-loop RNN are trained to predict random features of the future. This encourages the online observation representation to capture information that is useful to predict the future. This information is then distilled into the target observation encoder network through the EMA slow copy mechanism. In turn, these features become targets for the online network and predicting them can further improve the quality of the online representation. For further theoretical and empirical insights on why the bootstrap latent methods learn non-trivial representations see, e.g., \cite{tian2021understanding,wen2022mechanism}.

\subsection{Reward Normalization and Prioritization Scheme}

\paragraph{Reward Normalization.}
We use the world model uncertainties $\ell^{\,j}_t$ as an intrinsic reward.  To counter the non-stationarity  of the uncertainties during training, we adopt the same reward normalization scheme as~\RND~\cite{burda2019exploration} and divide the raw rewards $((\ell^{\,j}_t)_{t=0}^{T-2})_{j=0}^{B-1}$ by an EMA estimate of their standard deviation $\sigma_r$. The normalized rewards are $\ell^{\,j}_t/\sigma_r$. Details are provided in App.~\ref{app:rewardnormalization}.

\paragraph{Reward Prioritization.}
In addition to normalizing the rewards, we can optionally prioritize them by optimizing only the rewards with highest uncertainties and nullifying rewards with the lowest uncertainties. Because of the transient nature of the intrinsic rewards, this allows the agent to focus first on parts of the environment where the model is not accurate. Later on, if the previously nullified rewards remain, they will naturally become the ones with highest uncertainties and be optimized. This mechanism allows the agent to optimize only the source of high uncertainties and not optimize all sources of uncertainties at once. To do so, let us denote by $\mu_{\ell/\sigma_r}$ the adjusted EMA mean relative to the successive batch of normalized rewards $((\ell^{\,j}_t/ \sigma_r)_{t=0}^{T-2})_{j=0}^{B-1}$. We use $\mu_{\ell/\sigma_r}$ as a clipping threshold separating high and low-uncertainty rewards. Then, the clipped and normalized reward that plays the role of intrinsic reward is: $r^j_{i, t}=\max(\ell^{\,j}_t/\sigma_r-\mu_{\ell/\sigma_r}, 0)\cdot$
\subsection{Generic RL Algorithm and Representation Sharing}
\BYOLE can be used in conjunction with any RL algorithm for training the policy. In addition to providing an intrinsic reward, \BYOLE can further be used to shape the representation learnt by the RL agent by directly sharing some components of the \BYOLE world model with the RL model. For instance, consider a recurrent agent composed of an encoder $f_\psi$, an RNN cell $h^c_\psi$, a policy head $\pi_\psi$ and a value head $v_\psi$ that are shaped by an RL loss. Then, we can share the weights $\theta$ of the \BYOLE world model and the weights $\psi$ of the RL model at the level of the encoder and the RNN cell: $f_\psi=f_\theta$ and $h^c_\theta=h^c_\psi$ and let the joint representation be trained via both the RL loss and \BYOLE. In our experiments, we will show results for both the shared and unshared settings. Architectural details are provided in Appendix~\ref{app:byol_archi}.
\section{Experiments}
\label{sec:experiments}
We evaluate the algorithms on benchmark task-suites known to contain hard exploration challenges. These benchmarks have different properties in terms of the complexity of the observations, partial observability, and procedural generation, allowing us to test the generality of our approach. Our agents are implemented in the JAX ecosystem of libraries~\cite{deepmind2020jax}.

\label{subsec:environments}
\paragraph{Atari Learning Environment~\cite{bellemare2013arcade}.} This is a widely used RL benchmark, comprising of approximately $50$ \texttt{Atari} games. These are $2$-D, fully-observable, (fairly) deterministic environments for most of the games but have a very long optimization horizon (episodes last for an average of $10000$ steps) and complex observations (preprocessed greyscale images which are $84\times84$ byte arrays). We select the $10$ hardest exploration games~\cite{bellemare2016unifying} to conduct our experiments: \texttt{Alien}, \texttt{Freeway}, \texttt{Gravitar}, \texttt{Hero}, \texttt{Montezuma's Revenge}, \texttt{Pitfall}, \texttt{Private Eye}, \texttt{Qbert}, \texttt{Solaris} and \texttt{Venture}. 

\begin{figure}[htbp]
\centerline{
\includegraphics[width=.9\textwidth]{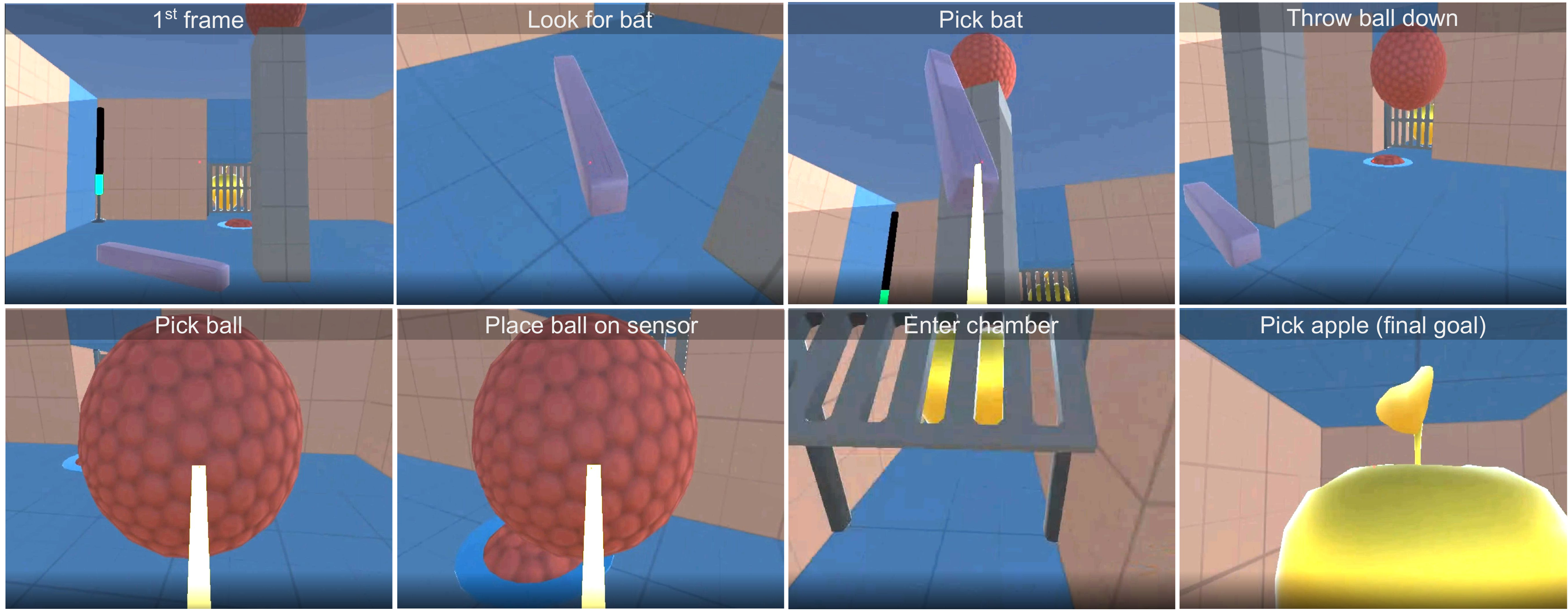}
}
\caption{$1^{\text{st}}$-person-view snapshots of the human player solving  \texttt{Baseball} task. They are  ordered chronologically from left to right and top to bottom. Each image depicts a specific stage of the task.}\label{fig:dmh_human_knock_obj_demo}
\end{figure}

\paragraph{Hard-Eight Suite~\cite{gulcehre2019making}.} This benchmark comprises of $8$ hard exploration tasks, originally built to emphasize the difficulties encountered by an RL agent when learning from sparse rewards in a procedurally-generated $3$-D world with partial observability, continuous control, and highly variable initial conditions. Each task requires the agent to interact with specific objects in its environment in order to reach a large apple that provides reward (see Fig.~\ref{fig:dmh_human_knock_obj_demo}). Being procedurally-generated, properties such as object shapes, colors, and positions are different every episode. The agent sees \textit{only} the first-person view from its position; but for greater clarity we provide \href{https://drive.google.com/drive/folders/1ERMPmP_UIqCNNhGgHO7Pu3IwX0ESQdn2?usp=sharing}{videos} showing top-down view and a third-person view as well to ground the difficulty of these tasks. Note that the current best RL agents that solve these tasks require a small (but non-zero) amount of human expert demonstrations. Without demonstrations or reward shaping, state-of-the-art deep RL algorithms, such as \RDD~\cite{kapturowski2018recurrent}, do not get positive reward signal on any of the tasks. In our case, we train a single RL agent and a single world model to tackle the 8 tasks all-together, making for a challenging multi-task setting.

\subsection{Experimental Setup}
At a high level, \BYOLE has $4$ main hyper-parameters: the target network EMA parameter $\alpha$, the open-loop horizon $K$, choosing to clip rewards and to share the \BYOLE representation with the RL network. To better understand what part of \BYOLE is essential to perform well, we run $4$ ablations. Each ablation corresponds to \BYOLE where only one hyper-parameter has been changed. The $4$ ablations are namely \textit{Fixed-targets} where the target network EMA parameter is set to $\alpha=1$, \textit{Horizon=1} where the horizon is set to $K=1$, \textit{No clipping} where we do not use clipping for the intrinsic rewards and \textit{No sharing} where we trained separately the RL network and the \BYOLE's world model.
In addition to \BYOLE, we also run as prominent baselines \RND, \ICM (see App.~\ref{app:baselines} for details), and pure RL which is an RL agent only using extrinsic rewards.

Finally, we run experiments on two different evaluation regimes. The first regime uses a mixed reward function $r_t=r_{e,t} + \lambda r_{i,t}$ which is a linear combination of the normalized extrinsic rewards $r_{e,t}$ and intrinsic rewards computed by the agent $r_{i,t}$ with mixing parameter $\lambda$. This may be the most important regime for a practitioner as we can see if our intrinsic rewards help improve performance, with respect to the extrinsic rewards, compared to the pure RL agent. The second regime is fully self-supervised where only the intrinsic reward $r_{i,t}$ is optimized. This regime gives us a sense of how pure exploration methods perform in complex environments.
\paragraph{Choice of RL algorithm.}
\label{subsec:RL choice}
We use VMPO~\cite{song2019v} as our RL algorithm. VMPO is an efficient on-policy optimization method that has achieved strong results across both discrete and continuous control tasks, and is thus applicable to all of the domains we consider. Further details regarding the RL algorithm setup and hyperparameters are provided in Appendix~\ref{app:vmpo_hypers}.
\paragraph{Performance Metrics.}
\label{subsec:results} We evaluate performance in terms of the agent score at learner step $t$, $\texttt{Agent}_\texttt{score}(t)$, as measured by undiscounted episode return.
We define the highest agent score through training as $\texttt{Agent}_\texttt{score}=\max_{t}\texttt{Agent}_\texttt{score}(t)$, as done in~\cite{fortunato2017noisy, badia2020agent57}. We define, for each game, the Human Normalized Score ($\texttt{HNS}$) at learner step $t$: $\texttt{HNS}(t) = \frac{\texttt{Agent}_\texttt{score}(t)-\texttt{Random}_\texttt{score}}{\texttt{Human}_\texttt{score}-\texttt{Random}_\texttt{score}}$ as well as the $\texttt{HNS}$ over the whole training: $\texttt{HNS} = \max_{t}\texttt{HNS}(t)$. A $\texttt{HNS}$ higher than $1$ means superhuman performance on a specific task. We similarly define the $\texttt{CHNS}$ Score as $\texttt{HNS}$ clipped between $0$ and $1$.

\begin{figure}[htbp]
    \centerline{
    \includegraphics[width=0.45\textwidth]{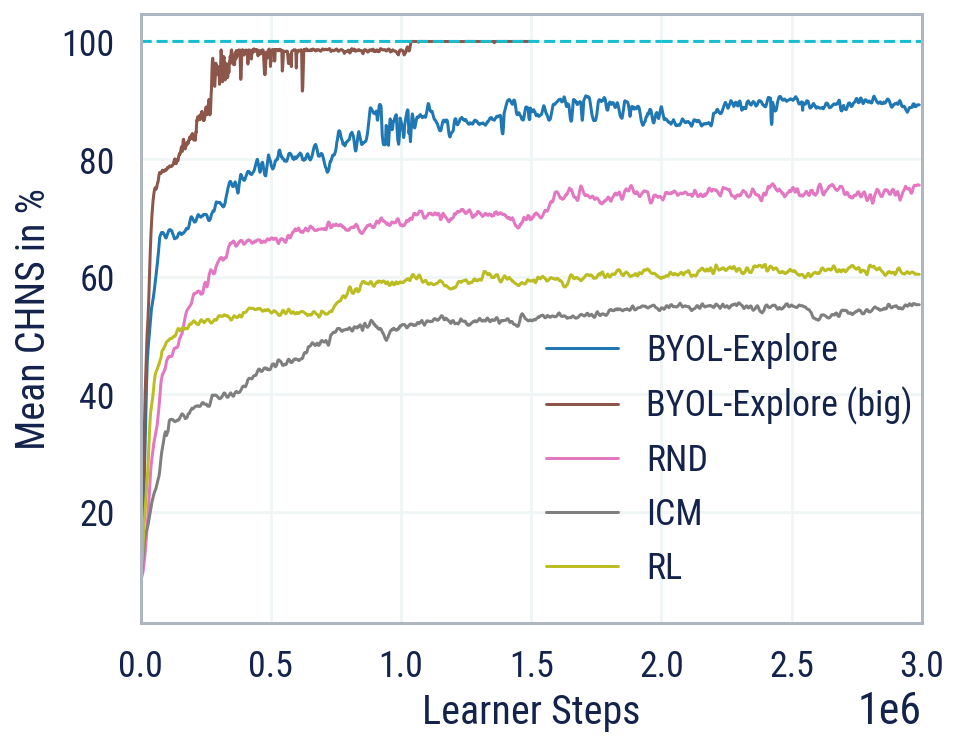}
    \hfill
    \includegraphics[width=0.45\textwidth]{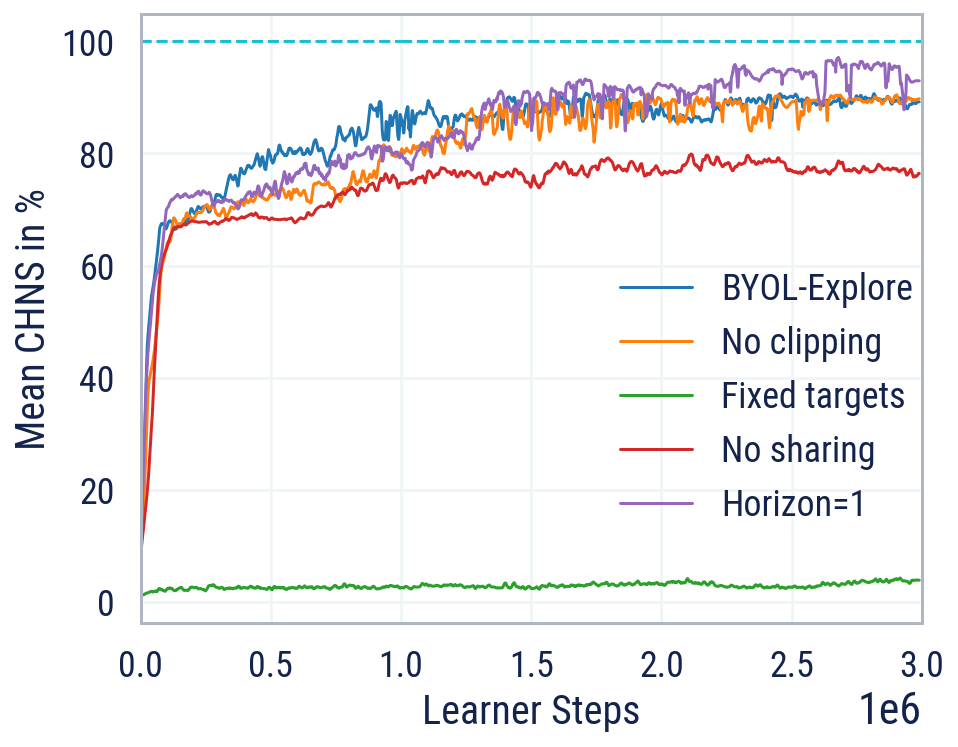}
    }
    \caption{Mean $\texttt{CHNS}(t)$ score across the tasks in $\texttt{Atari}$. \textbf{Left}:  \BYOLE and the baselines in the mixed regime for \texttt{Atari}. \textbf{Right}:  \BYOLE and its ablations in the mixed regime.}
    \label{fig:atari_chns_mixed_baselines}
\end{figure}

\subsection{Atari Results}

In these experiments, we set the target EMA rate $\alpha=0.99$ and open-loop horizon $K=8$.
We use $\lambda=0.1$ to combine the intrinsic and extrinsic rewards. We follow the classical $30$ random no-ops evaluation regime~\cite{mnih2015human, van2016deep}, and average performance over $10$ episodes and over $3$ seeds.

Fig.~\ref{fig:atari_chns_mixed_baselines}~(left) shows that \BYOLE is almost superhuman on the $10$-hardest exploration games and outperforms the different baselines of \RND, \ICM, and pure RL. Fig.~\ref{fig:atari_chns_mixed_baselines}~(right) compares \BYOLE against its ablations to gain finer insights into our method.
The \textit{No clipping} ablation performs comparably, showing that the prioritization of intrinsic rewards is not necessary on  \texttt{Atari} tasks.
Similarly, the \textit{Horizon=1} ablation performs slightly better, indicating that simply predicting one-step latents is sufficient to explore efficiently on the fully-observable \texttt{Atari} tasks.
The  \textit{Fixed Targets} ablation performs much worse, showing that our approach of predicting learned targets (rather than fixed random projections) is vital for good performance.
It is also worth noting that all the ablations except \textit{Fixed Targets} outperform all of our baselines, demonstrating the robustness of our approach.

Finally, because the \textit{Horizon=1} ablation was close to superhuman on \texttt{Atari}, we run the same configuration but double the length of the sequences on which we train from $64$ to $128$ (also doubling memory requirements while learning). For fair comparison, we train for only half the learner steps to keep total computation performed roughly equivalent.
With this small adjustment, this agent (\BYOLE~(big)) becomes superhuman on all of the $10$-hardest exploration games. 

\begin{wrapfigure}{r}{0.42\textwidth}
  \centering
  \includegraphics[width=0.42\textwidth]{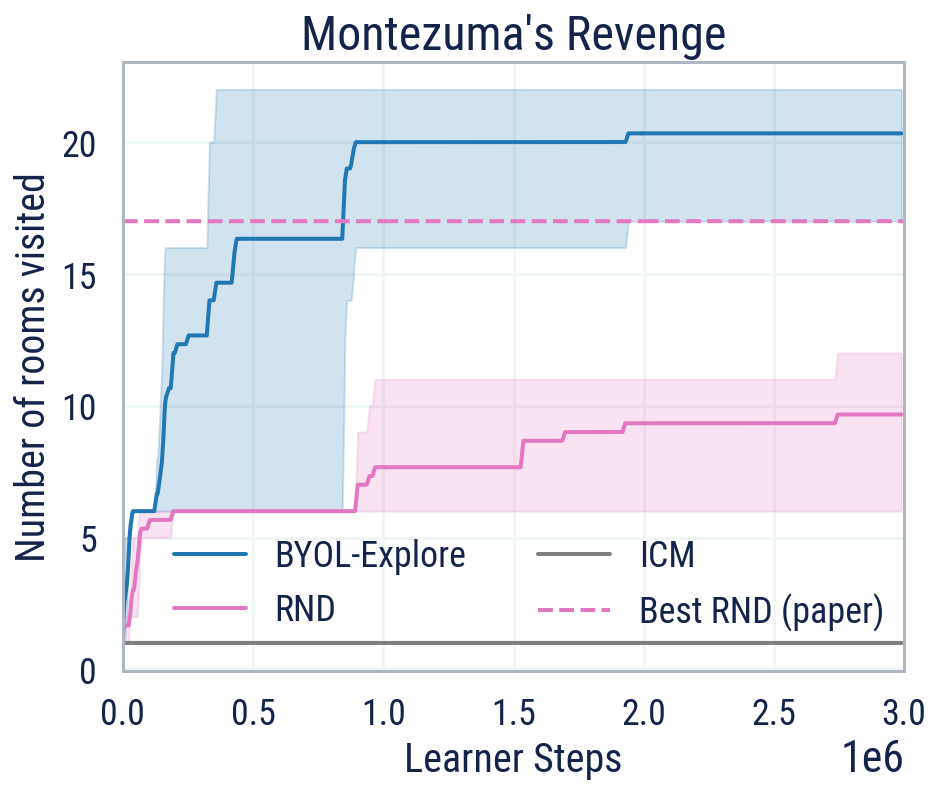}
  \caption{Number of rooms visited in \texttt{Montezuma's Revenge} during training in the self-supervised regime over $3$ seeds. \label{fig:atari_num_rooms}}
\end{wrapfigure}

\paragraph{Purely intrinsic exploration.} To test how \BYOLE behaves when only given intrinsic rewards without any extrinsic signal, we test on the well-known \texttt{Montezuma's Revenge} game by setting $\lambda = 0$. We measure exploratory behavior in terms of the number of different rooms of the dungeon the agent is able to explore over its lifetime. Note that accessing later rooms requires navigating complex dynamics such as collecting keys to open doors, avoiding enemies, and carefully traversing rooms filled with traps such as timed lasers. 
Figure \ref{fig:atari_num_rooms} shows how much room coverage is achieved during training when no extrinsic reward is used, showing that \BYOLE explores further than the best result reported by \RND~\cite{burda2019exploration}. Importantly, we use the episodic setting for intrinsic  rewards whereas the published \RND results considers the non-episodic setting for intrinsic rewards --- facilitating exploration as the agent is less risk-averse. Therefore, our setting could be considered even more challenging. Our agent explores more than $20$ rooms on average versus $17$ with best published \RND results. As expected in the episodic setting, our \RND re-implementation visits even fewer rooms. However, we can reproduce the published \RND results in the episodic setting when using recurrent policies.

\paragraph{Further results.} 
More fine-grained results are reported in App~\ref{app:experiments_atari_mixed}. We report, in Fig.\ref{fig:atari_curves_mixed_baselines} and in Fig.\ref{fig:atari_curves_mixed_ablations}, the agent scores learning curves for each game.  Tab.~\ref{tab:atari_score_mixed_baselines} and Tab.~\ref{tab:atari_score_mixed_ablations} have agent score at the end of training. Finally, Tab.~\ref{tab:stats_HNS_baselines} and Tab.~\ref{tab:stats_HNS_ablations} show the mean $\texttt{CHNS}$ and different statistics (mean and percentiles) of the $\texttt{HNS}$ across the selected games.

An interesting finding from examining the $\texttt{HNS}$ is that clipping and longer-horizon predictions are critical for very high scores on some games such as \texttt{Montezuma's Revenge} or \texttt{Hero}. \BYOLE has a median $\texttt{HNS}$ of $331.98$ compared to the \textit{No-clipping} ablation and the \textit{Horizon=1} which have a median $\texttt{HNS}$ of only $181.39$ and $199.80$ respectively. Therefore, while clipping is not necessary to get to human-level performance, it is still crucial to achieve top performance.

We also provide further results regarding the pure exploration setting on all 10 games in App.~\ref{app:experiments_atari_int}. 

\subsection{\DMH Results}

In these experiments, we set the target EMA rate $\alpha=0.99$ and open-loop horizon $K=10$.
We use $\lambda=0.01$ to combine the intrinsic and extrinsic rewards.

In contrast to prior work~\cite{gulcehre2019making}, we perform experiments in the more challenging multi-task regime, training a single agent to solve all eight tasks. At the beginning of each episode, a task is drawn uniformly at random from the suite.

\begin{figure}[htbp]
\centerline{
\includegraphics[width=.45\textwidth]{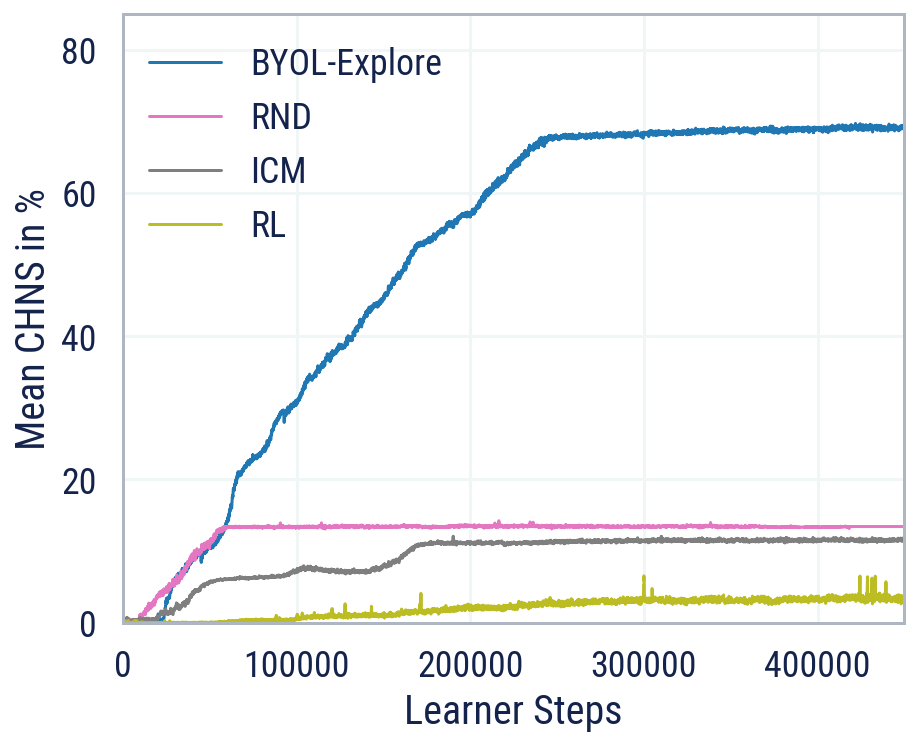}
\hfill
\includegraphics[width=.45\textwidth]{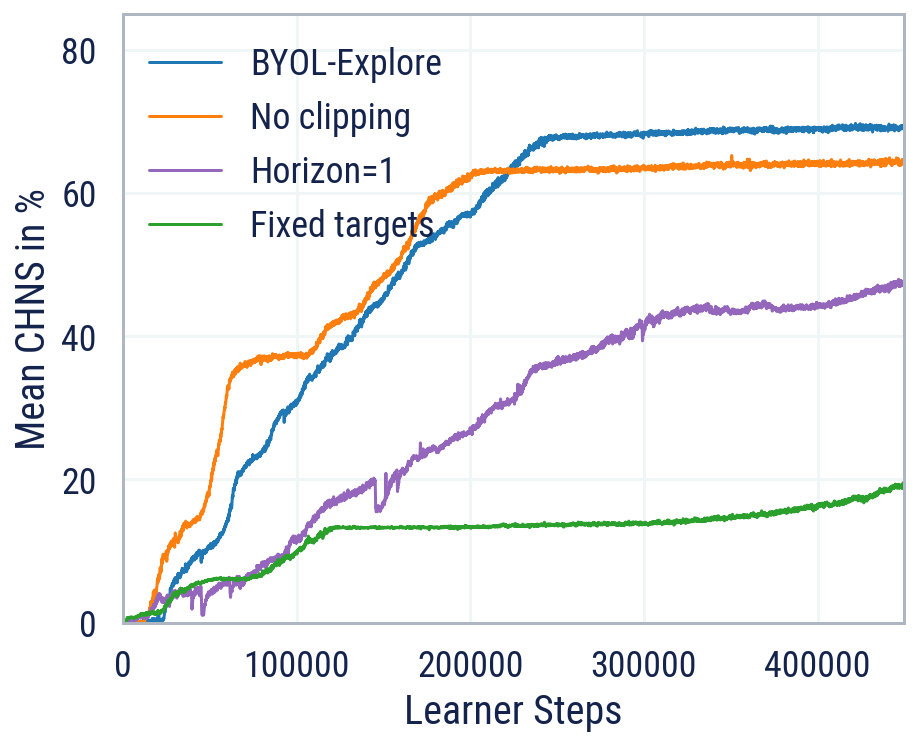}
}
\caption{Mean $\texttt{CHNS}(t)$ score across the tasks in the \DMH suite. \textbf{Left}: \BYOLE against baselines: \ICM, \RND and Pure RL. \textbf{Right}: \BYOLE against various ablations.}
\label{fig:dmh_chns_mixed_baselines}
\end{figure}

In Fig.~\ref{fig:dmh_chns_mixed_baselines}~(left) we report the mean $\texttt{CHNS}(t)$ across the tasks, averaged over 3 seeds. We see that \BYOLE outperforms the baselines of \RND, \ICM, and pure RL by a large margin.
Fig.~\ref{fig:dmh_chns_mixed_baselines}~(right) compares the performance of \BYOLE to its various ablations. Note that the \textit{No-clipping} ablation performs similarly to \BYOLE in terms of $\texttt{CHNS}$.
However, unlike the fully-observable \texttt{Atari} tasks, the \textit{Horizon=1} ablation learns considerably slower  and achieves lower final performance (see also our extended ablations on the horizon length in Fig.~\ref{fig:dmh_byole_horizon_ablation} in App.~\ref{app:hard_eight_experiments}).
We note once again that the \BYOLE bootstrapping mechanism for learning representations is essential, as confirmed by the poor (but non-zero) performance of the \textit{Fixed-targets} ablation.
Due to computational limitations, we did not run the \textit{No Sharing} ablation, as using separate networks requires twice the memory.

We now analyze our method more closely by examining per-task performance. The full learning curves for each task can be found in Fig.~\ref{fig:dmh_score_mixed_baselines} for \BYOLE and the main baselines and in Appendix~\ref{app:hard_eight_experiments} (see Fig.~\ref{fig:dmh_score_mixed_ablations}) for the various ablations. First, we take note that other curiosity-driven methods (\ICM and \RND) cannot get any positive score on the majority of the \DMH tasks, even with additional hyperparameter tuning and reward prioritizing (see Fig.~\ref{fig:dmh_tuned_baselines} and Fig.~\ref{fig:dmh_prioritized_baselines}  in App.~\ref{app:hard_eight_experiments}).

In contrast, we see that \BYOLE achieves strong performance on five out of the eight hard exploration tasks. Importantly, \BYOLE achieves this without human demonstrations, which was not the case in prior work ~\cite{gulcehre2019making}.
\BYOLE even surpasses humans on $4$ tasks, namely $\texttt{Navigate cubes}$, $\texttt{Throw-across}$, $\texttt{Baseball}$, and $\texttt{Wall Sensors}$ (see Tab.~\ref{tab:dmh_score_ablations} in App.~\ref{app:hard_eight_experiments} for details).
Most impressively, \BYOLE can solve $\texttt{Throw-across}$, which is a challenging task even for a skilful human player and was not solvable in prior work without collecting additional successful human demonstrations ~\cite{gulcehre2019making}.

Interestingly, note that on the $\texttt{Navigate Cubes}$ task, both \RND and the \textit{Fixed-targets} ablation achieve maximum performance alongside \BYOLE. We argue that this is because the prediction of random projections (either at the same step as done by \RND or multi-step as done by \BYOLE) leads to the policy learned performing spatial, navigational exploration --- this is the kind of behavior required to explore well on the $\texttt{Navigate Cubes}$ task. In contrast, the other tasks require exploratory behavior involving interaction with objects and the use of tools, where both \RND and the \textit{Fixed-targets} ablation fail.

Finally, we observe that two games, namely $\texttt{Remember Sensor}$ and $\texttt{Push Blocks}$, are particularly challenging, where all of our considered methods perform poorly.
We hypothesize that this is due to the larger variety of procedurally generated objects spawned in these levels, and the need to remember previous cues in the environment leading to a hard credit assignment problem.

\begin{figure}[htbp]
\includegraphics[width=.95\textwidth]{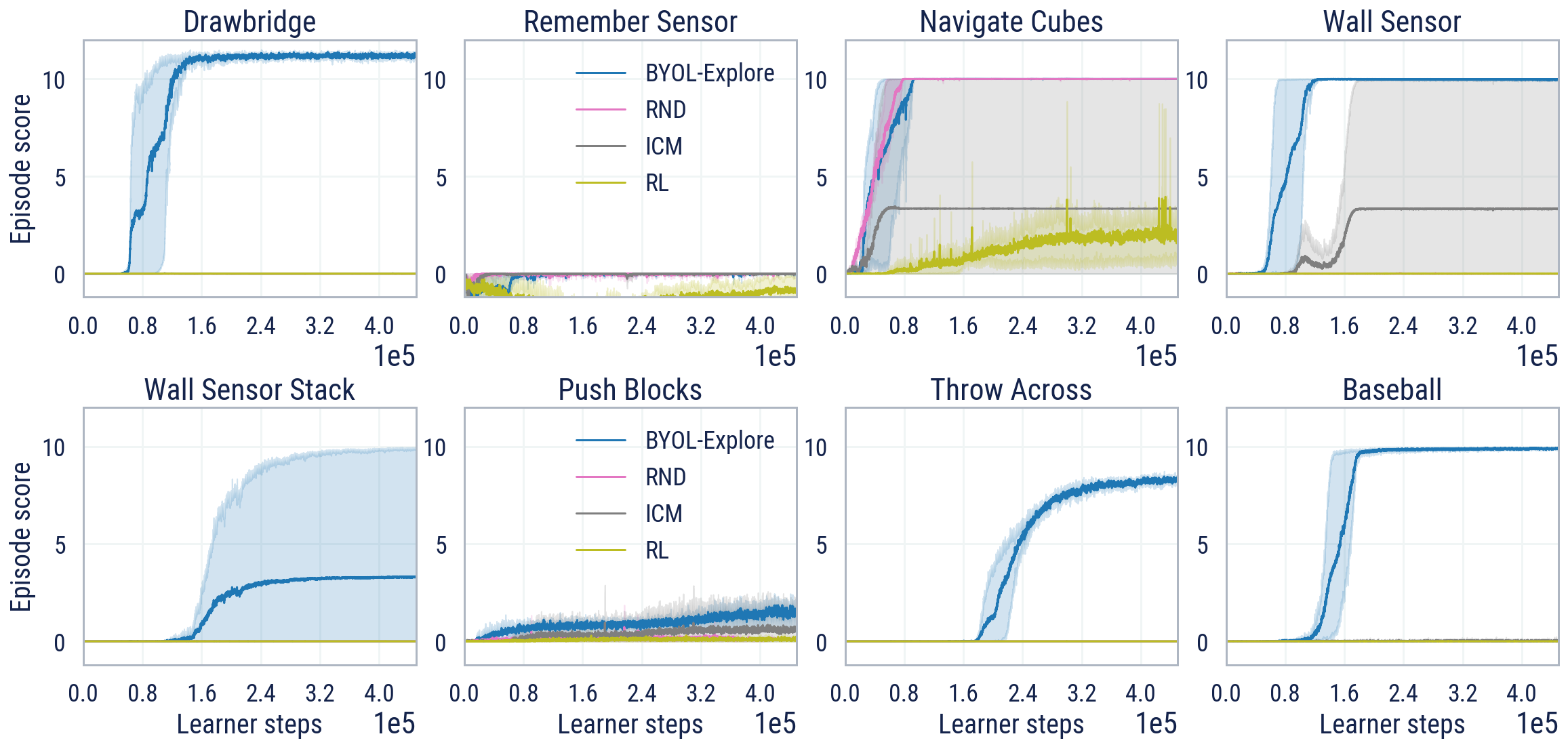}
\caption{Agent's score for each task in the \DMH suite  for \BYOLE against baselines. Shaded areas correspond to the minimum and maximum values across three seeds.}
\label{fig:dmh_score_mixed_baselines}
\end{figure}

\paragraph{Purely intrinsic exploration.} Each of the \DMH tasks has complex dynamics and object interactions, making it difficult to assess qualitatively the behavior of purely intrinsically motivated exploration. Nevertheless, for completeness, we provide results of \BYOLE trained only with intrinsic rewards in App.~\ref{app:hard_eight_experiments}, showing that it does achieve some positive signal on the \texttt{Drawbridge} and \texttt{Wall Sensor} tasks (see Fig.~\ref{fig:dmh_pure_intrinsic}).

\section{Conclusion}
\label{sec:conclusion}
We showed that \BYOLE is a simple curiosity-driven exploration method that achieves excellent performance on hard exploration tasks with fairly deterministic dynamics. \BYOLE is a multi-step prediction error method at the latent level that relies on recent advances in self-supervised learning to train its representation as well as its world-model without any additional loss. In~\texttt{Atari}, \BYOLE achieves superhuman performance on the $10$-hardest exploration games while being of much simpler design than other superhuman agents. Moreover, \BYOLE substantially outperforms previous exploration methods on \DMH navigation and manipulation tasks in a $3$-D, multi-task, partially-observable and procedurally-generated environment. This shows the generality of our algorithm to handle either $2$-D or $3$-D, single or multi-task, fully or partially-observable environments.

In the future, we would like to improve performance in \DMH and to demonstrate the generality of our method by extending it to other domains. In \DMH, we believe we can improve performance by scaling up the world model and finding better ways to trade off exploration and exploitation. Beyond \DMH, there are opportunities to tackle further challenges, most notably highly-stochastic and procedurally-generated environment dynamics such as \nethack~\cite{kuttler2020nethack}.
\clearpage

\begin{ack}
We would like to thank 
Abbas Abdolmaleki,
Arunkumar Byravan,
Adri\`a Puidomenech Badia,
Tim Harley,
Steven Kapturowski,
Thomas Keck, 
Jean-Baptiste Lespiau,
Kat McKinney,
Kyriacos Nikiforou, 
Georg Ostrovski,
Razvan Pascanu,
Doina Precup,
Satinder Singh,
Hubert Soyer,
Pablo Sprechmann,
and
Karl Tuyls
for their support and advice in developing and publishing this work.
\end{ack}

\newpage
\appendix

\section{General \BYOLE Architecture}
\label{app:byol_archi}
\begin{figure}[htbp]
    \centering
    \subfigure{\includegraphics[width=.8\textwidth]{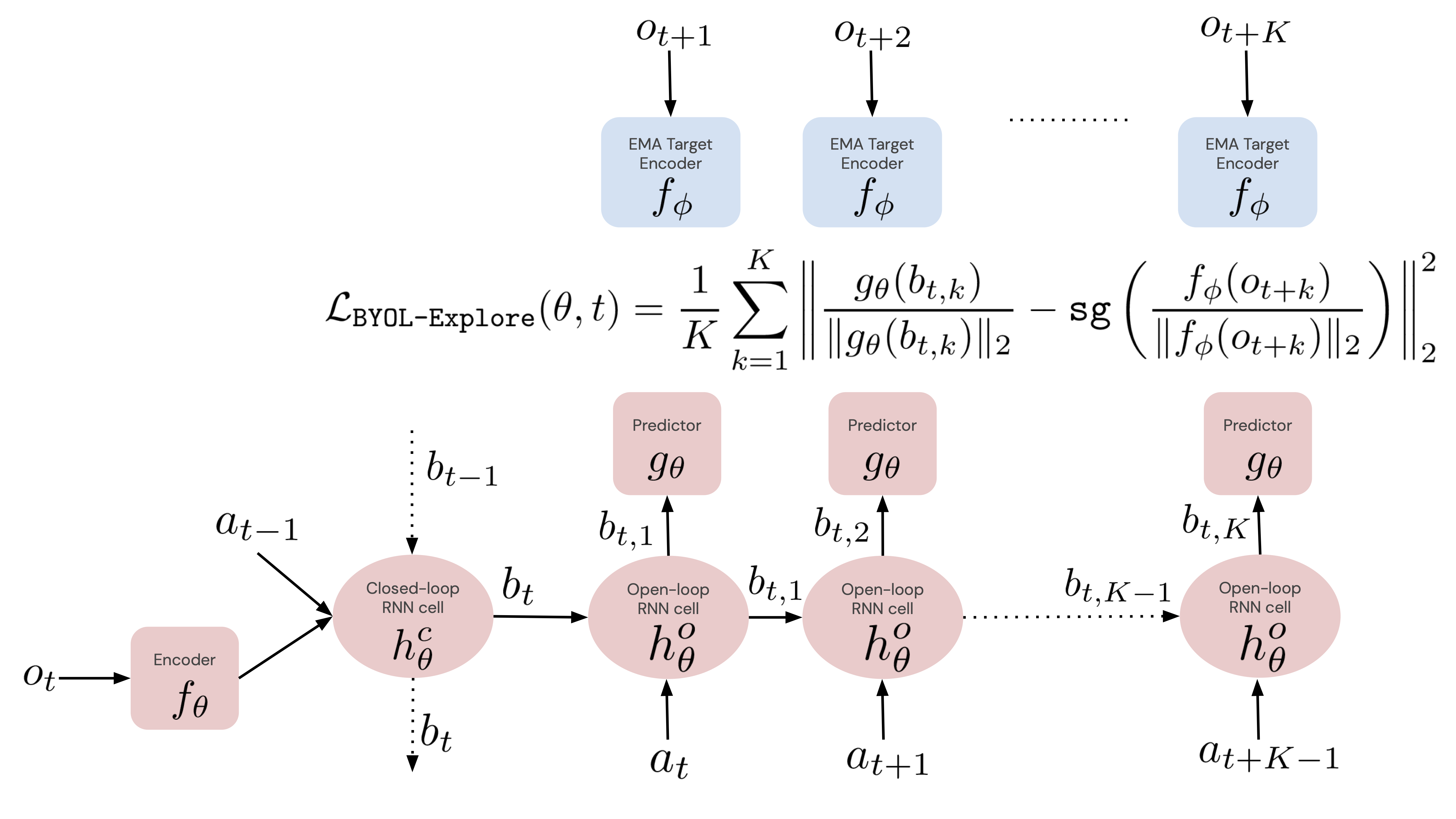}}
    \caption{\BYOLE's Neural Architecture.}
\end{figure}

The online network is composed of:
\begin{itemize}
    \item Encoder: $f_\theta: \mathcal{O}\rightarrow\mathbb{R}^N$
    \item Close-loop RNN cell: $h^c_\theta: \mathbb{R}^M\times\mathbb{R}^N\times\mathcal{A}\rightarrow\mathbb{R}^M$
    \item Open-loop RNN cell: $h^o_\theta: \mathbb{R}^M\times\mathcal{A}\rightarrow\mathbb{R}^M$
    \item Predictor: $g_\theta: \mathbb{R}^M\rightarrow\mathbb{R}^N$
\end{itemize}
The target network is composed of:
\begin{itemize}
    \item EMA encoder: $f_\phi: \mathcal{O}\rightarrow\mathbb{R}^N$
\end{itemize}

\subsection{Detailed \BYOLE architecture for \texttt{Atari}}

In \texttt{Atari}, the size of the observation-representation $N=512$ and the size of the history-representation $M=256$.

\begin{itemize}
    \item Encoder: $f_\theta:\mathcal{O}\rightarrow\mathbb{R}^N$: The encoder is instantiated as a Deep ResNet~\cite{he2016deep} stack. The greyscale image observation is passed through a stack of 3 units, each comprised of a $3\times 3$ convolutional layer, a $3 \times 3$ maxpool layer, and 2 residual blocks. The number of channels for the convolutional layer and the residual blocks are 16, 32, and 32 within each of the 3 units respectively. We use GroupNorm normalization~\cite{wu2018group} with one group at the end of each of the 3 units, and use ReLU activations everywhere. The output of the final residual block is flattened and projected using a single linear layer to an embedding of dimension $512$.
    \item Close-loop RNN cell: $h^c_\theta: \mathbb{R}^M\times\mathbb{R}^N\times\mathcal{A}\rightarrow\mathbb{R}^M$ is a simple Gated Recurrent Unit (GRU)~\cite{cho2014properties}. We provide the past-action to the close-loop RNN cell, embedded into a representation of size $32$.
    \item Open-loop RNN cell: $h^o_\theta: \mathbb{R}^M\times\mathcal{A}\rightarrow\mathbb{R}^M$ is a simple Gated Recurrent Unit. We provide the past-action to the open-loop RNN cell, embedded into a representation of size $32$.
    \item Policy head $\pi_\psi : \mathbb{R}^N \rightarrow \mathbb{R}^{|\mathcal{A}|}$, value head $v_\psi : \mathbb{R}^N \rightarrow \mathbb{R}$, and \BYOLE predictor $g_\theta : \mathbb{R}^M\rightarrow\mathbb{R}^N$ are each a simple Multi-Layer Perceptron (MLP) with one hidden layer of size $256$. The outputs of the policy head are passed through a softmax layer to form the probabilities for each action to be taken.
\end{itemize}

\subsection{Detailed \BYOLE architecture for \DMH}

In \DMH, the size of the observation-representation $N=256$ and the size of the history-representation $M=512$.

\begin{itemize}
    \item Encoder: $f_\theta:\mathcal{O}\rightarrow\mathbb{R}^N$: The encoder operates over RGB image observations at each timestep, and is implemented as a Deep ResNet~\cite{he2016deep} stack. The image is passed through a stack of 3 units, each comprised of a $3\times 3$ convolutional layer with stride 1, a $3 \times 3$ maxpool layer with stride 2, and 2 residual blocks. We use ReLU activations everywhere. The output of this stack is flattened and passed through a linear layer of width 1024, followed by LayerNorm and ReLU layers, and finally a linear layer of width 256.
    \item Close-loop RNN cell: $h^c_\theta: \mathbb{R}^M\times\mathbb{R}^N\times\mathcal{A}\rightarrow\mathbb{R}^M$: The RNN cell operates over the output of the encoder $f_\theta$ in addition to some other observation signals per timestep, which we describe first. The observation contains (i) the force the agent's hand is currently applying as a fraction of total grip strength, a single scalar between 0 and 1; (ii) a boolean indicating whether the hand is currently holding an object; (iii) the distance of the agent hand from the main body; (iv) the last action taken by the agent; each of these are embedded using a linear projection to 20 dimensions followed by a ReLU. (v) the previous reward obtained by the agent, passed by a signed hyperbolic transformation; (vi) a text instruction specific to the task currently being solved, with each word embedded into 20 dimensions and processed sequentially by an LSTM to an embedding of size 64. All of these quantities, along with the output of $f_\theta$, are concatenated and passed through a linear layer to an embedding of size 512. An LSTM with embedding size 512 processes each of these observation representations sequentially to form the recurrent history-representation. 
    \item Open-loop RNN cell: $h^o_\theta: \mathbb{R}^M\times\mathcal{A}\rightarrow\mathbb{R}^M$: This is implemented as an LSTM. We discretize the action into 76 bins and provide them as one-hot representations of the action for the partial history unroll.
    \item Policy head $\pi_\psi : \mathbb{R}^N \rightarrow \mathbb{R}^{|\mathcal{A}|}$, value head $v_\psi : \mathbb{R}^N \rightarrow \mathbb{R}$, and \BYOLE predictor $g_\theta: \mathbb{R}^M\rightarrow\mathbb{R}^N$: These are each implemented as an MLP with hidden sizes of (512,), (512,), and (128, 256, 512,) respectively. The policy network uses different linear projections of the shared hidden layer to compute components of the policy over different parts of the action space. The action space has a mix of both discrete (modeled using a softmax layer of logits computed as linear projection of the hidden layer) as well as continuous (modeled as Gaussian distributions over each dimension with the mean and variance modeled using a linear projection of the hidden layer) actions, described in detail in prior works~\cite{gulcehre2019making}.
\end{itemize}

\subsection{Detailes of Reward Normalization Mechanism}
\label{app:rewardnormalization}

We use a similar reward normalization scheme as in~\RND~\cite{burda2019exploration} and normalize the raw rewards $((\ell^{\,j}_t)_{t=0}^{T-2})_{j=0}^{B-1}$ by an EMA estimate of their standard deviation.

More precisely, we first set the EMA mean to $\overline{r}=0$, the EMA mean of squares to $\overline{r^2}=0$ and the counter to $c=1$. Then, for the $c$-th batch  of raw rewards $((\ell^{\,j}_t)_{t=0}^{T-2})_{j=0}^{B-1}$, we compute the batch mean $\overline{r_c}$ and the batch mean of squares $\overline{r^2_c}$ :
\begin{equation*}
\overline{r_c} = \frac{1}{B(T-1)}\sum_{j=0}^{B-1}\sum_{t=0}^{T-2}\ell^{\,i}_t,\qquad \overline{r^2_c} = \frac{1}{B(T-1)}\sum_{j=0}^{B-1}\sum_{t=0}^{T-2}(\ell^{\,i}_t)^2.\\
\end{equation*}
We then update $\overline{r}$, $\overline{r^2}$ and $c$:
\begin{equation*}
\overline{r} \leftarrow \alpha_r\overline{r}+ (1-\alpha_r)\overline{r_c},\qquad \overline{r^2} \leftarrow \alpha_r\overline{r^2}+ (1-\alpha_r)\overline{r^2_c},\qquad c\leftarrow c+1,
\end{equation*}
where $\alpha_r=0.99$. We compute the adjusted EMA mean $\mu_r$, the adjusted EMA mean of squares $\mu_{r^2}$:
\begin{equation*}
    \mu_r = \frac{\overline{r}}{1-\alpha_r^c}\CommaBin \qquad \mu_{r^2} = \frac{\overline{r^2}}{1-\alpha_r^c}\cdot
\end{equation*}
Finally the EMA estimation of the standard deviation is $\sigma_r = \sqrt{\max(\mu_{r^2} - \mu_r^2,0)+\epsilon}$,
where $\varepsilon = 10^{-8}$ is a small numerical regularization. The normalized rewards are $\ell^{\,j}_t/\sigma_r$.

\section{Baselines}
\label{app:baselines}
\paragraph{Random Network Distillation (\RND)~\cite{burda2019exploration}} is a simple exploration method that consists in training an encoder such that its outputs fit the outputs of \textit{another} fixed and randomly initialized encoder and using the training loss as an intrinsic reward to be optimized by an RL algorithm. More precisely, let $N\in\mathbb{N}^*$ be the embedding size and let us note $f_\theta:\mathcal{O}\rightarrow\mathbb{R}^N$ the encoder, also called predictor network, with trainable weights $\theta$ and $f_\phi:\mathcal{O}\rightarrow\mathbb{R}^N$ the fixed and randomly initialized encoder, also called target network, with fixed weights $\phi$. In addition, let us suppose that we have a batch of trajectories $\left((o^j_t, a^j_t)_{t=0}^{T-1}\right)_{j=0}^{B-1}$ collected by our RL agent, then the loss $\mathcal{L}_{\texttt{RND}}(\theta)$ to minimize w.r.t. the online network parameters is defined as:
\begin{equation*}
\mathcal{L}_{\texttt{RND}}(\theta,j,t)= \|f_\theta(o^j_{t})-\texttt{sg}(f_\phi(o^j_{t}))\|_2^2,\quad \mathcal{L}_{\texttt{RND}}(\theta) = \frac{1}{BT}\sum_{j=0}^{B-1}\sum_{t=0}^{T-1}\mathcal{L}_{\texttt{RND}}(\theta,j,t),    
\end{equation*}
and the unnormalized reward associated to the transition $(o^j_t, a^j_t, o^j_{t+1})$ is defined as $\ell^{\,j}_t=\mathcal{L}_{\texttt{RND}}(\theta,j,t+1)$ where $0\leq t\leq T-2$. To obtained the final intrinsic rewards, we just normalize them to be as close as possible to the original \RND implementation:$r^j_{i, t}=\frac{\ell^{\,i}_t}{\sigma_r}$.

\begin{figure}[htbp]
    \centering
    \subfigure{\includegraphics[width=.4\textwidth]{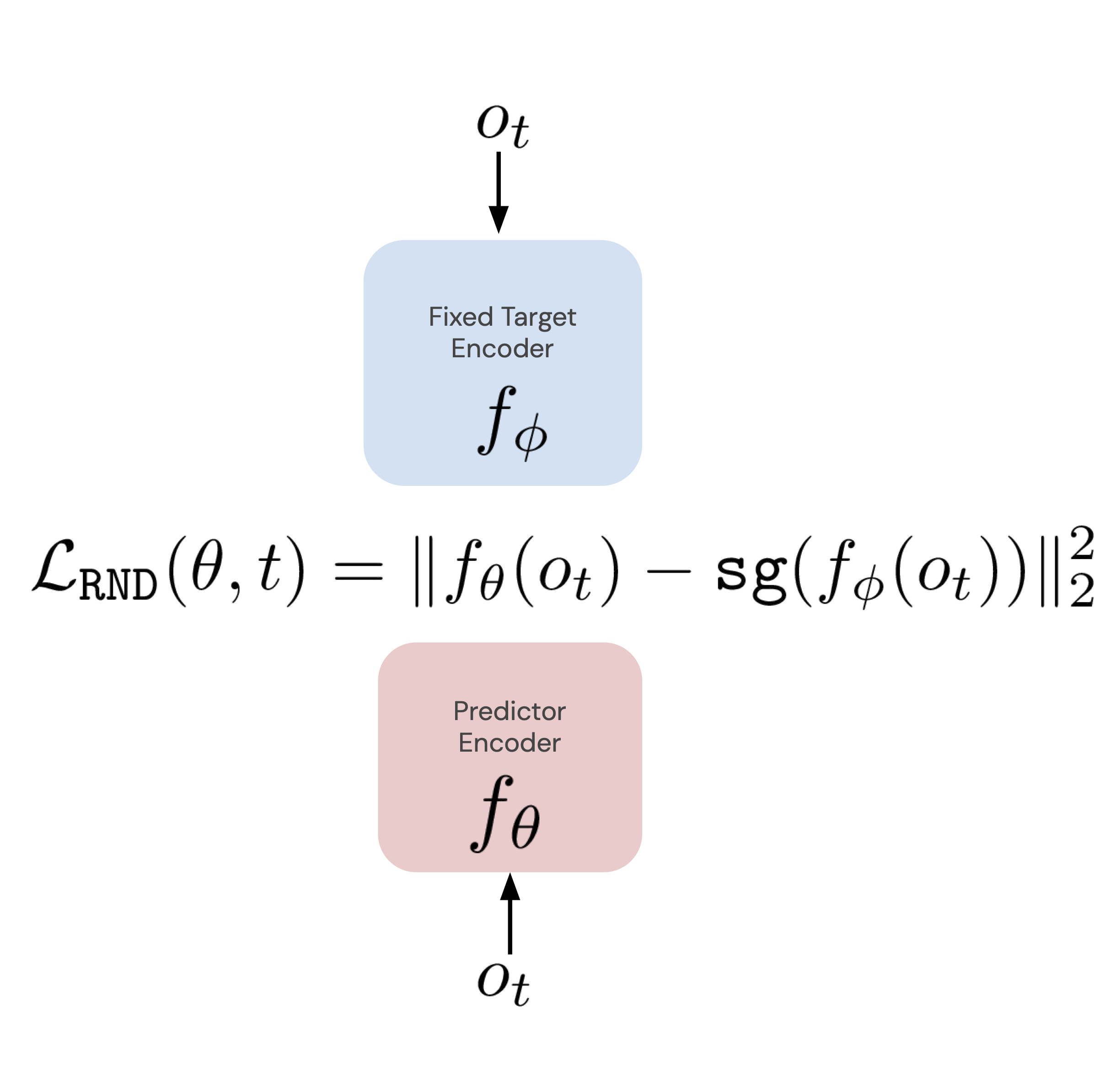}}
    \caption{\RND's Neural Architecture.}
\end{figure}

\paragraph{Intrinsic Curiosity Module (\ICM)~\cite{pathak2017curiosity}} is a one-step prediction error method at the latent level. It consists in training an encoder $f_\theta:\mathcal{O}\rightarrow\mathbb{R}^N$ that outputs a representation that is robust to uncontrollable aspects of the environment and then use this representation as inputs of a one-step prediction error model $g_\phi:\mathbb{R}^N\times\mathcal{A}\rightarrow\mathbb{R}^N$ which error is used as an intrinsic reward to be optimized by an RL algorithm. To build a representation robust to uncontrollable dynamics, the idea used in \ICM is to train an inverse dynamics model $p_\theta:\mathbb{R}^N\times\mathcal{R}^N\rightarrow\mathcal{A}$ that predicts the distribution of actions that led to the transition between two consecutive representations $f_\theta(o_t), f_\theta(o_{t+1})$. More precisely, let us suppose that we have a batch of trajectories $\left((o^j_t, a^j_t)_{t=0}^{T-1}\right)_{j=0}^{B-1}$ collected by our RL agent, then the loss $\mathcal{L}_{\texttt{INV}}(\theta)$ to minimize in order to train our encoder and inverse dynamcis model is:
\begin{equation*}
\mathcal{L}_{\texttt{INV}}(\theta,j,t)= - \ln\left(p_\theta(a^j_t|f_\theta(o^j_t), f_\theta(o^j_{t+1}))\right),\quad\mathcal{L}_{\texttt{INV}}(\theta) = \frac{1}{B(T-1)}\sum_{j=0}^{B-1}\sum_{t=0}^{T-2}\mathcal{L}_{\texttt{INV}}(\theta,j,t),    
\end{equation*}
which is a simple cross-entropy loss. Simultaneously, \ICM also trains the one step prediction error model by minimizing the following one-step prediction loss:
\begin{equation*}
\mathcal{L}_{\ICM}(\phi,j,t)= \|g_\phi(f_\theta(o^j_t), a^j_t)-\texttt{sg}(f_\theta(o^j_{t+1}))\|_2^2, \quad \mathcal{L}_{\ICM}(\phi) = \frac{1}{B(T-1)}\sum_{j=0}^{B-1}\sum_{t=0}^{T-2}\mathcal{L}_{\ICM}(\theta,j,t),    
\end{equation*}
and the unnormalized reward associated to the transition $(o^j_t, a^j_t, o^j_{t+1})$ is defined as $\ell^{\,j}_t=\mathcal{L}_{\ICM}(\theta,j,t)$ where $0\leq t\leq T-2$. To obtained the final intrinsic rewards, we just normalize them :$r^j_{i, t}=\frac{\ell^{\,j}_t}{\sigma_r}$.

\begin{figure}[htbp]
    \centering
    \subfigure{\includegraphics[width=.9\textwidth]{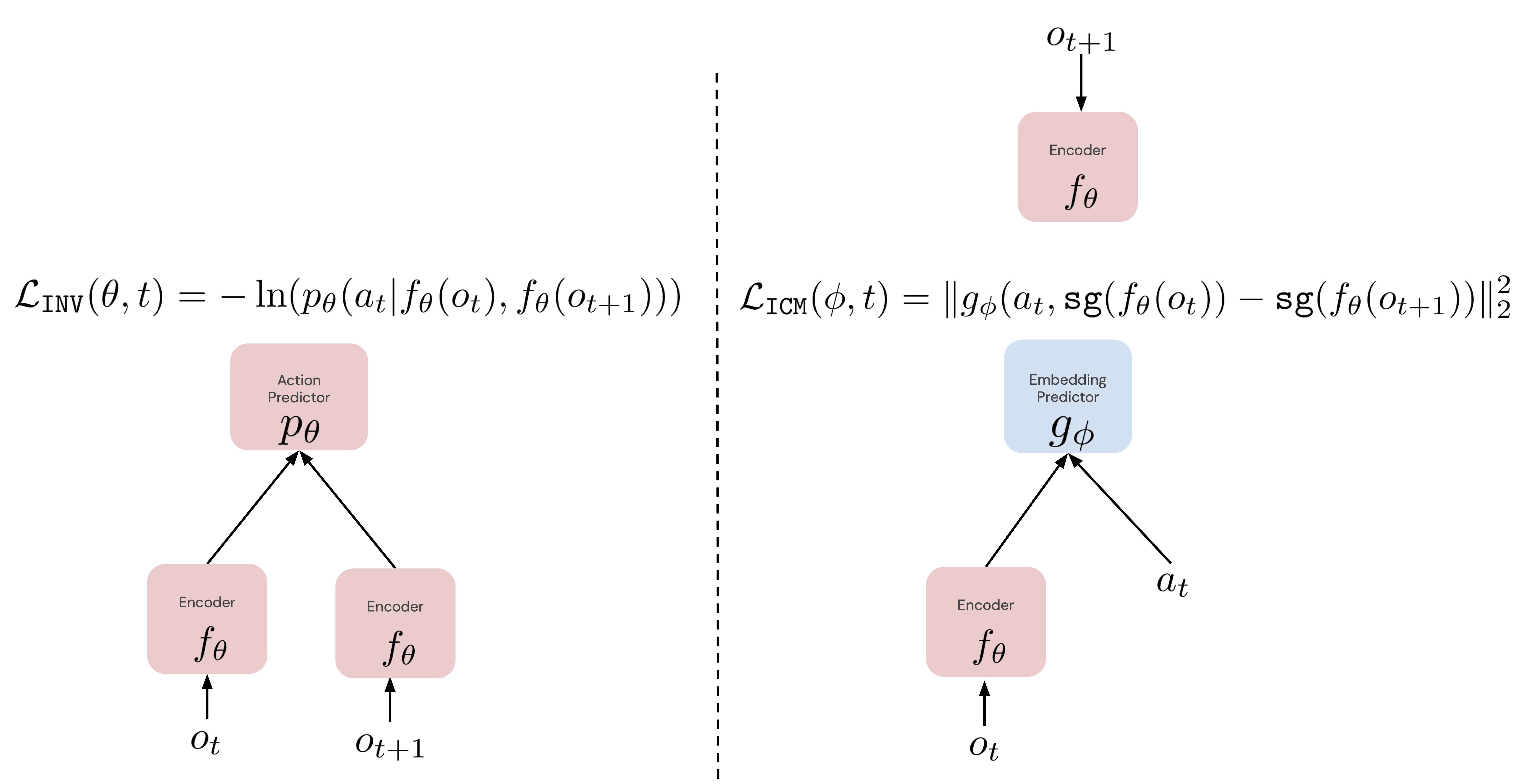}}
    \caption{\ICM's Neural Architecture.}
\end{figure}

\newpage

\section{Hyperparameter Settings}
\label{app:vmpo_hypers}

\paragraph{\texttt{Atari} Settings}
We perform PopArt-style~\cite{hessel2019multi} reward normalization with a step size of 0.01, and after normalizing rescale the rewards by $1 - \gamma$. We also similarly use PopArt normalization on the output of the value network.

We use a discount factor of $\gamma = 0.999$. To train the value function, we use VTrace without offpolicy corrections to define TD targets for MSE loss with a loss weight of 0.5. We add an entropy loss with a loss weight of 0.001. The VMPO parameters $\eta_{init}$ and $\alpha_{init}$ are initialized to 0.5. $\epsilon_{\eta}$ and $\epsilon_{\alpha}$ are set to 0.01 and 0.005 respectively. We scale the BYOL loss by a factor of 5.0 when combining losses. The VMPO top-k parameter is set to 0.5.
We use the Adam optimizer with learning rate $10^{-4}$ and $b_1=0.9$. The target network for VMPO is updated every 10 learner steps.

We use a batch size of 32 and a sequence length of 128; and a distributed learning setup using 4 TPUv2 for learning and 400 CPU actors for generating data via another inference server using 4 TPUv2 to evaluate the policy, similar to Agent57~\cite{badia2020agent57}.

\paragraph{\DMH Settings}
We perform separately PopArt for both the intrinsic and extrinsic rewards, on the multi-task suite of environments. We use a step size of $10^{-5}$ for extrinsic popart and 0.01 for intrinsic popart.

We use a discount factor of $\gamma = 0.99$.
To train the value function, we use VTrace without offpolicy corrections to define TD targets for MSE loss with a loss weight of 0.5. We do not add an entropy loss term.
We scale the BYOL loss by a factor of 1.0 when combining losses. The VMPO top-k parameter is set to 0.5.
The VMPO temperature parameters $\eta_{init}$ and $\epsilon_{\eta}$ are set to 1.0 and 0.1 respectively.
For the discrete components of the action, the VMPO KL constraint parameters $\alpha_{init\_categorical}$ and $\epsilon_{\alpha\_categorical}$ are set to 5.0 and 0.0016 respectively. 
For the continuous Gaussian actions, we found it important to apply the VMPO contraints separately to the mean and covariance of the distributions, per action dimension.
Thus for the mean, $\alpha_{init\_mean}$ and $\epsilon_{\alpha\_mean}$ are 1.0 and 0.001 respectively; 
for the covariance, $\alpha_{init\_covariance}$ and $\epsilon_{\alpha\_covariance}$ are 1.0 and 0.0001 respectively.

We use the Adam optimizer with learning rate $10^{-4}$ and $b_1=0.0$. The target network for VMPO is updated every 10 learner steps.

We use a batch size of 48 and a sequence length of 80; and use a distributed learning setup consisting of 8 TPUv3 divided into 6 for learning and 2 for acting similar to the Sebulba~\cite{hessel2021podracer} framework.

\section{Further Experimental Results}
\label{app:experiments}
\subsection{\texttt{Atari}: Detailed results when mixing intrinsic and extrinsic rewards}
\label{app:experiments_atari_mixed}

For more detailed analysis of our agent performance, we provide detailed results broken down per-task and by showing statistics of the human normalized score across games other than simply the capped mean.

\begin{figure}[htbp]
    \centering
    \includegraphics[width=1.0\textwidth]{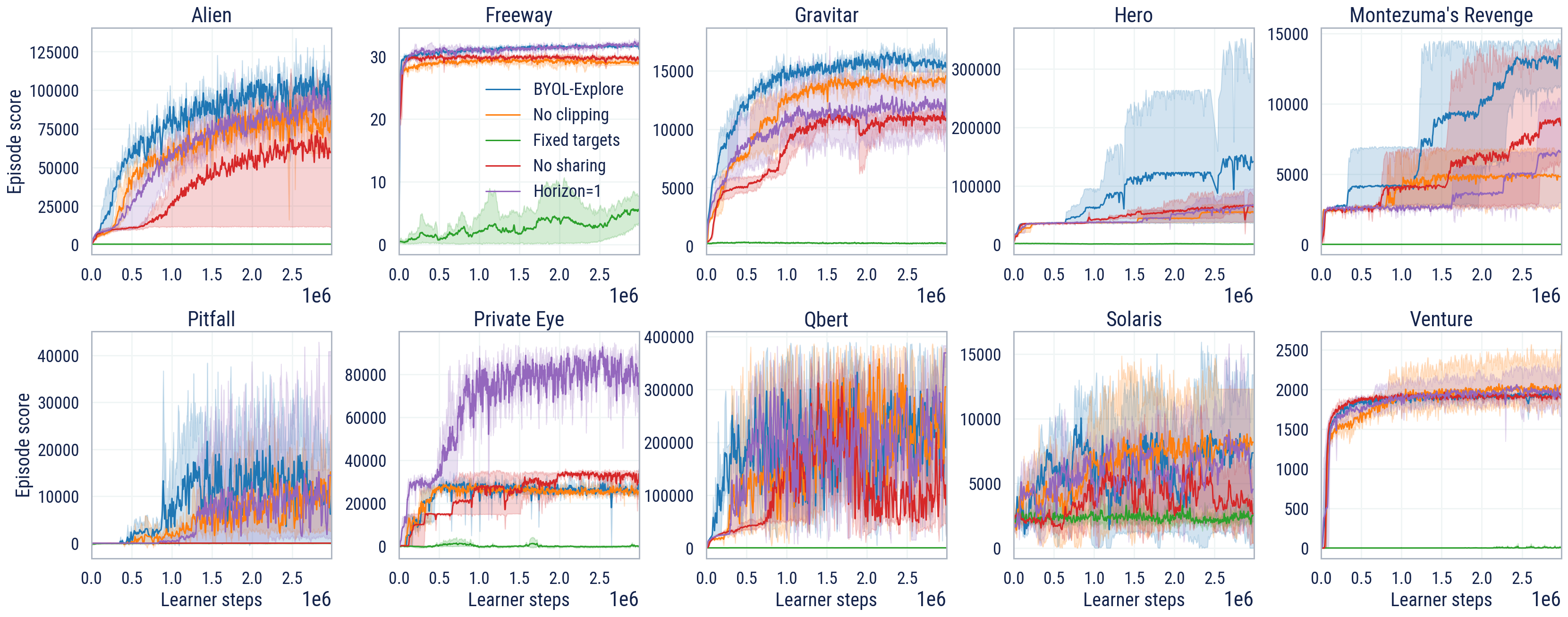}
    \caption{Learning curves in terms of agent score for \BYOLE and baselines for \texttt{Atari}, averaged over 10 episodes and 3 seeds.}
    \label{fig:atari_curves_mixed_baselines}
\end{figure}

\begin{figure}[htbp]
    \centering
    \includegraphics[width=1.0\textwidth]{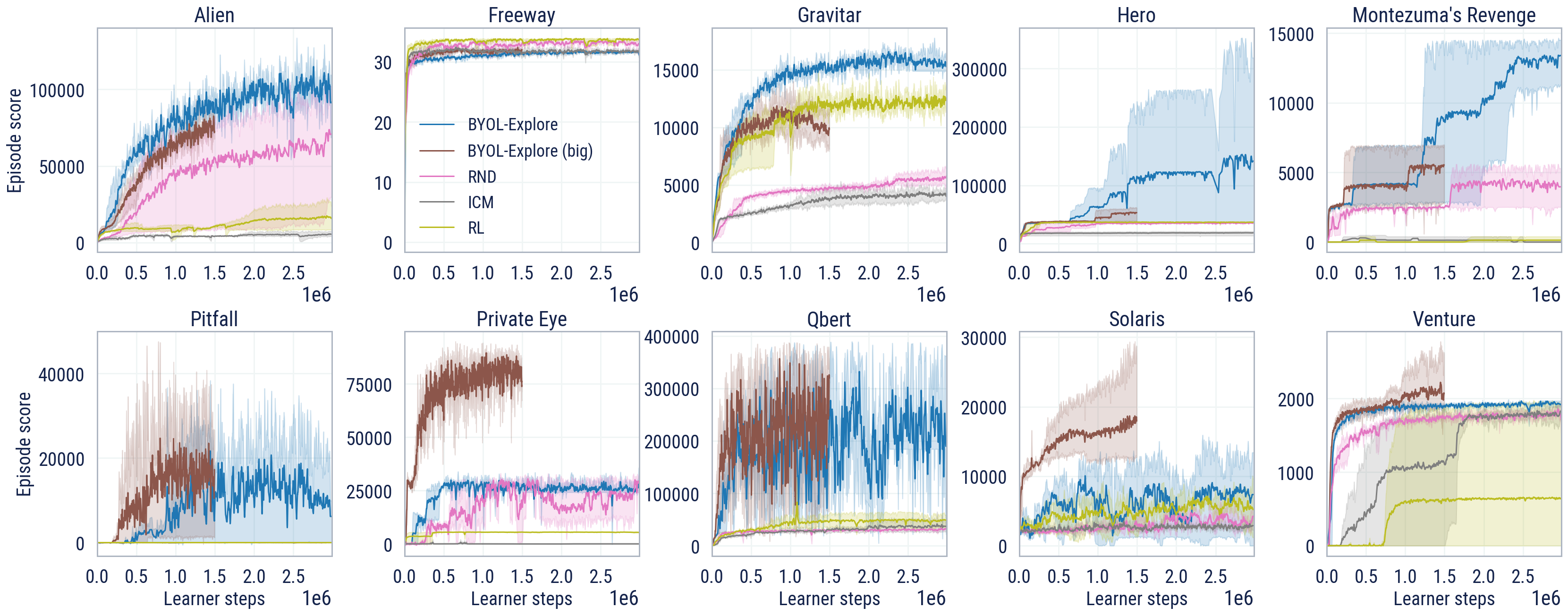}
    \caption{Learning curves in terms of agent score for \BYOLE and its ablations for \texttt{Atari}, averaged over 10 episodes and 3 seeds.}
    \label{fig:atari_curves_mixed_ablations}
\end{figure}

Table~\ref{tab:atari_score_mixed_baselines} and Figure~\ref{fig:atari_curves_mixed_baselines} compare agent performance for \BYOLE and our various baselines considered, while Table~\ref{tab:atari_score_mixed_baselines} and Figure~\ref{fig:atari_curves_mixed_ablations} compare \BYOLE to its various ablations.

Tables~\ref{tab:stats_HNS_baselines}~and~\ref{tab:stats_HNS_ablations} show various finer-grained statistics of human normalized score averaged across all games, for \BYOLE compared to our baselines and its ablations, respectively.

\begin{table}[htbp]
\begin{adjustwidth}{-2cm}{-2cm}
\small
\centering
\begin{tabular}{|c|c|c|c|c|c|c|}
\hline
 Games & Average Human & BYOL-Explore & RND & ICM & RL & BYOL-Explore (big) \\
\hline
 alien & 7127.70 & \bf{124524.02} & 74717.02 & 5664.83 & 18252.69 & 87754.33 \\
 freeway & 29.60 & 32.12 & 33.74 & 32.61 & \bf{33.97} & 32.62 \\
 gravitar & 3351.40 & \bf{17014.51} & 5885.23 & 4657.20 & 13541.10 & 12234.71 \\
 hero & 30826.40 & \bf{153124.29} & 36452.78 & 18753.97 & 37235.84 & 54139.73 \\
 montezuma revenge & 4753.30 & \bf{13518.18} & 4554.37 & 300.00 & 160.11 & 5615.36 \\
 pitfall & 6463.70 & 25175.53 & 0.00 & 0.00 & 0.00 & \bf{27555.12} \\
 private eye & 69571.30 & 33130.85 & 33150.09 & 1419.49 & 5704.99 & \bf{94589.47} \\
 qbert & 13455.00 & 358269.62 & 34341.06 & 44294.63 & 91920.56 & \bf{377349.65} \\
 solaris & 12326.70 & 11060.24 & 5262.55 & 3924.22 & 8147.67 & \bf{19277.35} \\
 venture & 1187.50 & 1980.24 & 1879.68 & 1860.19 & 670.61 & \bf{2250.12} \\
\hline
\end{tabular}
\captionsetup{margin={1cm,-3cm},justification=centering}
\caption{Maximum agent score obtained over training for \BYOLE and the baselines for 10 hard-exploration \texttt{Atari} games, averaged over 10 episodes and 3 seeds.}
\label{tab:atari_score_mixed_baselines}
\normalsize
\end{adjustwidth}
\end{table}

\begin{table}[htbp]
\begin{adjustwidth}{-2cm}{-2cm}
\small
\centering
\begin{tabular}{|c|c|c|c|c|c|c|}
\hline
 Games & Average Human & BYOL-Explore & No clipping & Fixed targets & No sharing & Horizon=1 \\
\hline
 alien & 7127.70 & \bf{124524.02} & 99570.34 & 411.63 & 75876.95 & 105384.32 \\
 freeway & 29.60 & 32.12 & 30.13 & 6.09 & 30.63 & \bf{32.73} \\
 gravitar & 3351.40 & \bf{17014.51} & 14949.85 & 401.79 & 12026.79 & 13621.41 \\
 hero & 30826.40 & \bf{153124.29} & 55897.06 & 1963.10 & 67555.39 & 67954.99 \\
 montezuma revenge & 4753.30 & \bf{13518.18} & 5093.32 & 8.52 & 9090.71 & 6701.77 \\
 pitfall & 6463.70 & \bf{25175.53} & 19427.77 & -3.55 & 0.00 & 17363.46 \\
 private eye & 69571.30 & 33130.85 & 31048.56 & 1706.72 & 36183.65 & \bf{95359.23} \\
 qbert & 13455.00 & 358269.62 & \bf{370719.60} & 565.59 & 338709.48 & 368757.50 \\
 solaris & \bf{12326.70} & 11060.24 & 10380.94 & 3772.82 & 7105.23 & 9585.28 \\
 venture & 1187.50 & 1980.24 & \bf{2121.38} & 31.12 & 1977.89 & 2078.24 \\
\hline
\end{tabular}
\captionsetup{margin={1cm,-3cm},justification=centering}
\caption{Maximum agent score obtained over training for \BYOLE and the ablations for 10 hard-exploration \texttt{Atari} games, averaged over 10 episodes and 3 seeds.}
\label{tab:atari_score_mixed_ablations}
\normalsize
\end{adjustwidth}
\end{table}

\begin{table}[ht]
\small
\centering
\begin{tabular}{|c|c|c|c|c|c|}
\hline
 Statistics & BYOL-Explore & RND & ICM & RL & BYOL-Explore (big) \\
\hline
 Mean CHNS & 93.62 & 78.32 & 57.43 & 63.38 & \bf{100.00} \\
 Number of superhuman games & 8 & 6 & 4 & 5 & \bf{10} \\
 Mean HNS & \bf{661.14} & 209.08 & 91.42 & 174.22 & 579.57 \\
 Median HNS & \bf{331.98} & 116.44 & 69.14 & 88.55 & 183.86 \\
 40\% Percentile HNS & \bf{237.34} & 106.72 & 45.39 & 59.98 & 172.01 \\
 30\% Percentile HNS & 149.28 & 81.36 & 18.86 & 41.98 & \bf{154.66} \\
 20\% Percentile HNS & 104.52 & 45.37 & 5.73 & 7.22 & \bf{132.41} \\
 10\% Percentile HNS & 84.48 & 33.02 & 3.29 & 3.42 & \bf{117.34} \\
 5\% Percentile HNS & 66.04 & 18.22 & 2.65 & 3.39 & \bf{113.78} \\
\hline
\end{tabular}
\caption{Statistics of human-normalized score (HNS) and Clipped HNS (CHNS) over the $10$-hardest exploration games for \BYOLE and baselines, averaged over 10 episodes and 3 seeds.}
\label{tab:stats_HNS_baselines}
\normalsize
\end{table}

\begin{table}[ht]
\small
\centering
\begin{tabular}{|c|c|c|c|c|c|}
\hline
 Statistics & BYOL-Explore & No clipping & Fixed targets & No sharing & Horizon=1 \\
\hline
 Mean CHNS & 93.62 & 92.71 & 6.81 & 80.83 & \bf{97.53} \\
 Number of superhuman games & 8 & 8 & 0 & 7 & \bf{9} \\
 Mean HNS & \bf{661.14} & 568.52 & 6.81 & 480.94 & 584.68 \\
 Median HNS & \bf{331.98} & 181.39 & 3.08 & 178.90 & 199.80 \\
 40\% Percentile HNS & \bf{237.34} & 150.05 & 2.88 & 141.32 & 161.40 \\
 30\% Percentile HNS & \bf{149.28} & 105.54 & 2.65 & 88.30 & 139.82 \\
 20\% Percentile HNS & 104.52 & 97.92 & 2.58 & 52.73 & \bf{131.78} \\
 10\% Percentile HNS & 84.48 & 78.67 & 2.19 & 47.14 & \bf{107.05} \\
 5\% Percentile HNS & 66.04 & 61.64 & 1.19 & 25.28 & \bf{91.17} \\
\hline
\end{tabular}
\caption{Statistics of human-normalized score (HNS) and Clipped HNS (CHNS) over the $10$-hardest exploration games for \BYOLE and its ablations, averaged over 10 episodes and 3 seeds.}
\label{tab:stats_HNS_ablations}
\normalsize
\end{table}

\subsection{\texttt{Atari}: Detailed results for the pure exploration regime}
\label{app:experiments_atari_int}

To further evaluate \BYOLE as an approach for curiousity-driven exploration, we examine the performance of an agent trained only using intrinsic rewards, without ever using the extrinsic reward signal. In the task \texttt{Montezuma's Revenge}, Figure~\ref{fig:atari_num_rooms} measures the quality of exploraton in terms of the number of different rooms of the dungeon the agent is able to explore without using extrinsic rewards. However, computing this requires access to privileged information and detailed understanding of the environment design, and this kind of analysis is not easily adapted to a wider range of diverse games. Thus, as a proxy for engaging in interesting behavior in the environment, we measure exploratory behavior in terms of the extrinsic reward the agent obtains. As obtaining large amounts of extrinsic reward requires visiting many different regions of the state space and efficient exploration is highly correlated with this behavior, this is a reasonable measure for evaluating the quality of our agents. However, we stress again that though we measure performance in terms of the extrinsic reward, the agent is never given access to these reward signals while training.

\begin{figure}[htbp]
    \centering
    \includegraphics[width=1.0\textwidth]{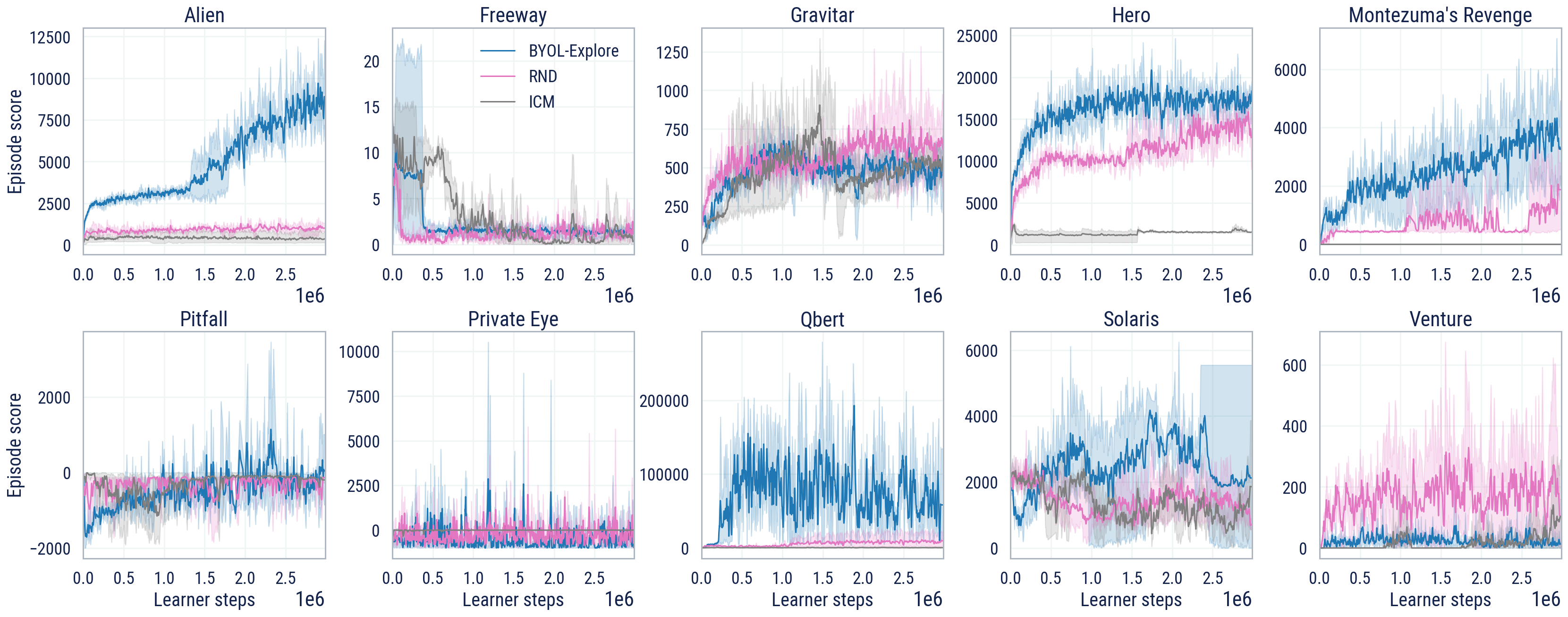}
    \caption{Learning curves in terms of agent score for \BYOLE and its ablations in the pure-exploration regime for \texttt{Atari}, averaged over 10 episodes and 3 seeds.}
    \label{fig:atari_curves_int_baselines}
\end{figure}

Table~\ref{tab:atari_score_int_baselines} shows a summary of performance in each of the 10 hard-exploration games we consider, while Figure~\ref{fig:atari_curves_int_baselines} shows performance through training.

Notice that \BYOLE is the only method to get positive rewards in \texttt{Pitfall}, showing that it can learn to effectively navigate the complex environment even without complex mechanisms such as Episodic Memory Modules used by prior works~\cite{badia2020agent57}.
 
\begin{table}[ht!]
\small
\centering
\begin{tabular}{|c|c|c|c|c|}
\hline
 Games & Average Human & BYOL-Explore & RND & ICM \\
\hline
 alien & 7127.70 & \bf{10964.95} & 1441.69 & 691.90 \\
 freeway & \bf{29.60} & 12.94 & 9.89 & 15.74 \\
 gravitar & \bf{3351.40} & 795.95 & 954.84 & 1040.45 \\
 hero & \bf{30826.40} & 22078.89 & 16964.23 & 2567.09 \\
 montezuma revenge & 4753.30 & \bf{5146.73} & 2305.65 & 2.68 \\
 pitfall & \bf{6463.70} & 1498.92 & -83.55 & -7.53 \\
 private eye & \bf{69571.30} & 5495.88 & 4309.31 & 469.09 \\
 qbert & 13455.00 & \bf{200081.13} & 12233.18 & 1062.49 \\
 solaris & \bf{12326.70} & 4298.20 & 2896.69 & 2851.86 \\
 venture & \bf{1187.50} & 101.25 & 399.53 & 157.99 \\
\hline
\end{tabular}  
\caption{Maximum agent score obtained through training for \BYOLE and the baselines in the pure-exploration regime for \texttt{Atari}, averaged over 10 episodes and 3 seeds.}
\label{tab:atari_score_int_baselines}
\normalsize
\end{table}

\subsection{\DMH Experiments}
\label{app:hard_eight_experiments}

We similarly examine various statistics of agent performance on the \DMH suite in terms of human-normalized score for \BYOLE versus our baselines (in Figure~\ref{tab:dmh_chns_baselines}) and its ablations (in Figure~\ref{tab:dmh_chns_ablations}).

\begin{table}[ht!]
\small
\centering
\begin{tabular}{|c|c|c|c|c|}
\hline
 Statistics & BYOL-Explore & RND & ICM & RL \\
\hline
 Mean CHNS & \bf{69.87} & 14.36 & 12.42 & 7.10 \\
 Number of superhuman games & \bf{4} & 1 & 0 & 0 \\
 Mean HNS & \bf{83.92} & 17.89 & 12.42 & 7.10 \\
 Median HNS & \bf{101.21} & 0.21 & 3.35 & 0.21 \\
 40\% Percentile HNS & \bf{81.74} & 0.09 & 0.83 & 0.15 \\
 30\% Percentile HNS & \bf{43.98} & 0.02 & 0.24 & 0.11 \\
 20\% Percentile HNS & \bf{28.71} & 0.00 & 0.13 & 0.05 \\
 10\% Percentile HNS & \bf{17.20} & 0.00 & 0.08 & 0.01 \\
 5\% Percentile HNS & \bf{11.45} & 0.00 & 0.04 & 0.01 \\
\hline
\end{tabular}
\caption{Statistics of human-normalized score (HNS) and Clipped HNS (CHNS) over the \DMH suite for \BYOLE and baselines, averaged over 3 seeds.}
\label{tab:dmh_chns_baselines}
\normalsize
\end{table}

\begin{table}[ht!]
\small
\centering
\begin{tabular}{|c|c|c|c|c|}
\hline
 Statistics & BYOL-Explore & No clipping & Fixed targets & Horizon=1 \\
\hline
 Mean CHNS & \bf{69.87} & 65.45 & 19.85 & 50.60 \\
 Number of superhuman games & \bf{4} & 4 & 1 & 1 \\
 Mean HNS & \bf{83.92} & 79.50 & 23.38 & 51.32 \\
 Median HNS & \bf{101.21} & 100.44 & 2.91 & 65.03 \\
 40\% Percentile HNS & \bf{81.74} & 78.15 & 0.10 & 52.30 \\
 30\% Percentile HNS & \bf{43.98} & 33.18 & 0.09 & 17.50 \\
 20\% Percentile HNS & \bf{28.71} & 14.10 & 0.05 & 8.18 \\
 10\% Percentile HNS & \bf{17.20} & 4.02 & 0.01 & 3.79 \\
 5\% Percentile HNS & \bf{11.45} & 2.10 & 0.01 & 2.04 \\
\hline
\end{tabular}
\caption{Statistics of human-normalized score (HNS) and Clipped HNS (CHNS) over the \DMH suite for \BYOLE and its ablations, averaged over 3 seeds.}
\label{tab:dmh_chns_ablations}
\normalsize
\end{table}

Next, we examine per-task performance across each task within the \DMH suite for \BYOLE compared with our baselines (in Table~\ref{tab:dmh_score_baselines} and Figure~\ref{fig:dmh_score_mixed_baselines}), and compared with our main ablations (in Table~\ref{tab:dmh_score_ablations}).

\begin{table}[ht!]
\small
\centering
\begin{tabular}{|c|c|c|c|c|c|}
\hline
 Games & Average Human & BYOL-Explore & RND & ICM & RL \\
\hline
 Drawbridge & \bf{12.30} & 11.38 & 0.00 & 0.02 & 0.00 \\
 Remember Sensor & \bf{7.60} & 0.00 & 0.00 & 0.00 & -0.29 \\
 Navigate Cubes & 7.80 & \bf{10.00} & 10.00 & 3.46 & 3.95 \\
 Wall Sensor & 9.10 & \bf{10.00} & 0.03 & 3.33 & 0.02 \\
 Wall Sensor Stack & \bf{8.60} & 3.32 & 0.00 & 0.01 & 0.01 \\
 Push Blocks & \bf{8.40} & 1.86 & 0.73 & 0.96 & 0.30 \\
 Throw Across & 5.70 & \bf{8.46} & 0.00 & 0.00 & 0.00 \\
 Baseball & 7.90 & \bf{9.94} & 0.01 & 0.08 & 0.01 \\
\hline
\end{tabular}
\caption{Maximum agent score obtained through training for \BYOLE and our baselines for the various tasks in the \DMH suite, averaged over 3 seeds.}
\label{tab:dmh_score_baselines}
\normalsize
\end{table}

\begin{table}[ht!]
\small
\centering
\begin{tabular}{|c|c|c|c|c|c|}
\hline
 Games & Average Human & BYOL-Explore & No clipping & Fixed targets & Horizon=1 \\
\hline
 Drawbridge & \bf{12.30} & 11.38 & 11.19 & 0.00 & 7.66 \\
 Remember Sensor & \bf{7.60} & 0.00 & -0.00 & 0.00 & -0.03 \\
 Navigate Cubes & 7.80 & \bf{10.00} & 10.00 & 10.00 & 8.25 \\
 Wall Sensor & 9.10 & \bf{10.00} & 10.00 & 4.24 & 6.66 \\
 Wall Sensor Stack & \bf{8.60} & 3.32 & 0.02 & 0.01 & 0.02 \\
 Push Blocks & \bf{8.40} & 1.86 & 2.25 & 0.52 & 1.05 \\
 Throw Across & 5.70 & 8.46 & \bf{8.48} & 0.00 & 3.87 \\
 Baseball & 7.90 & \bf{9.94} & 9.91 & 0.01 & 6.59 \\
\hline
\end{tabular}
\caption{Maximum agent score obtained through training for \BYOLE and its main ablations for the various tasks in the \DMH suite, averaged over 3 seeds.}
\label{tab:dmh_score_ablations}
\normalsize
\end{table}

\begin{figure}[tb]
\includegraphics[width=1.\textwidth]{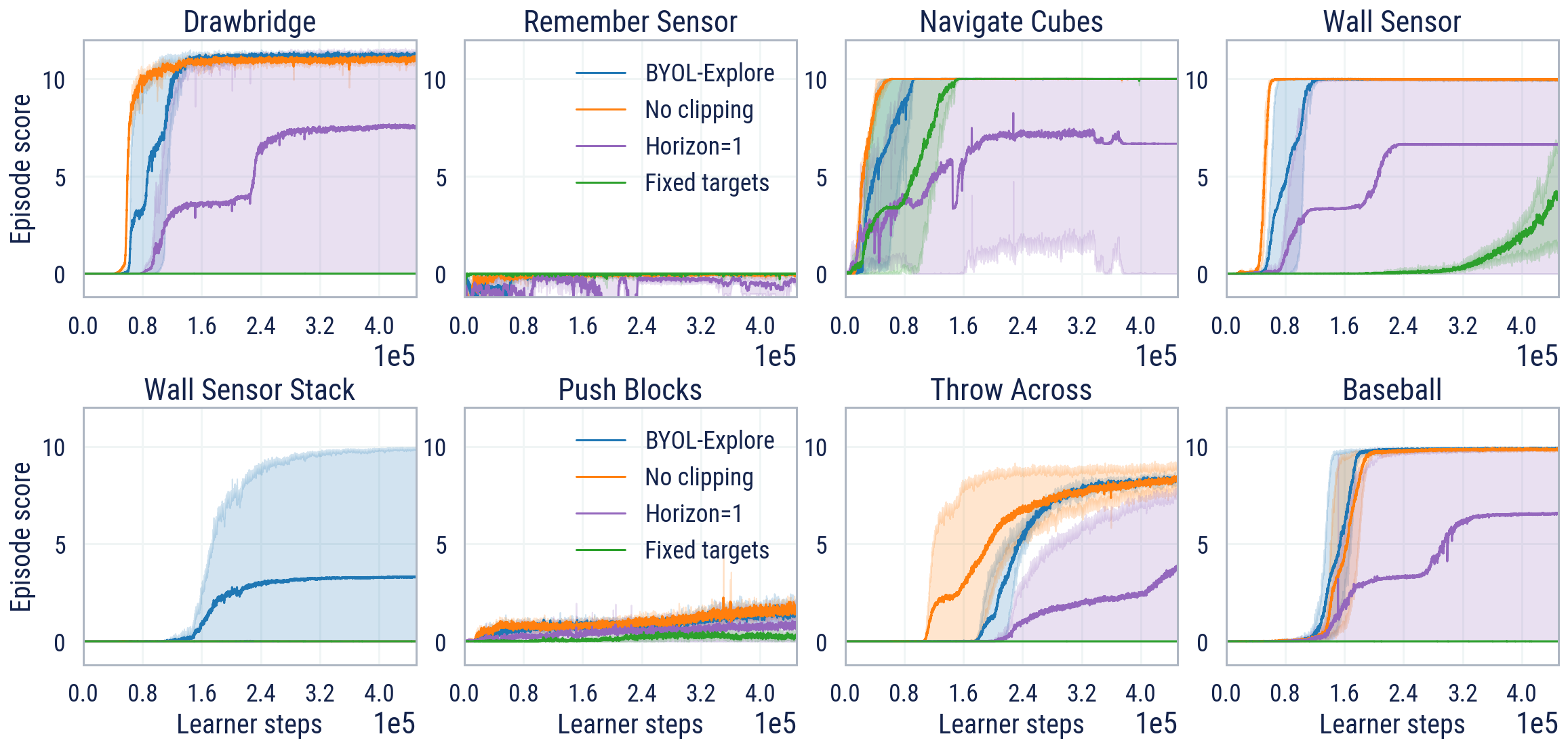}
\caption{Learning curves in terms of agent score for \BYOLE and its main ablations for each task in the \DMH suite, averaged over 3 seeds.}
\label{fig:dmh_score_mixed_ablations}
\end{figure}

\begin{figure}[tb]
\includegraphics[width=1.\textwidth]{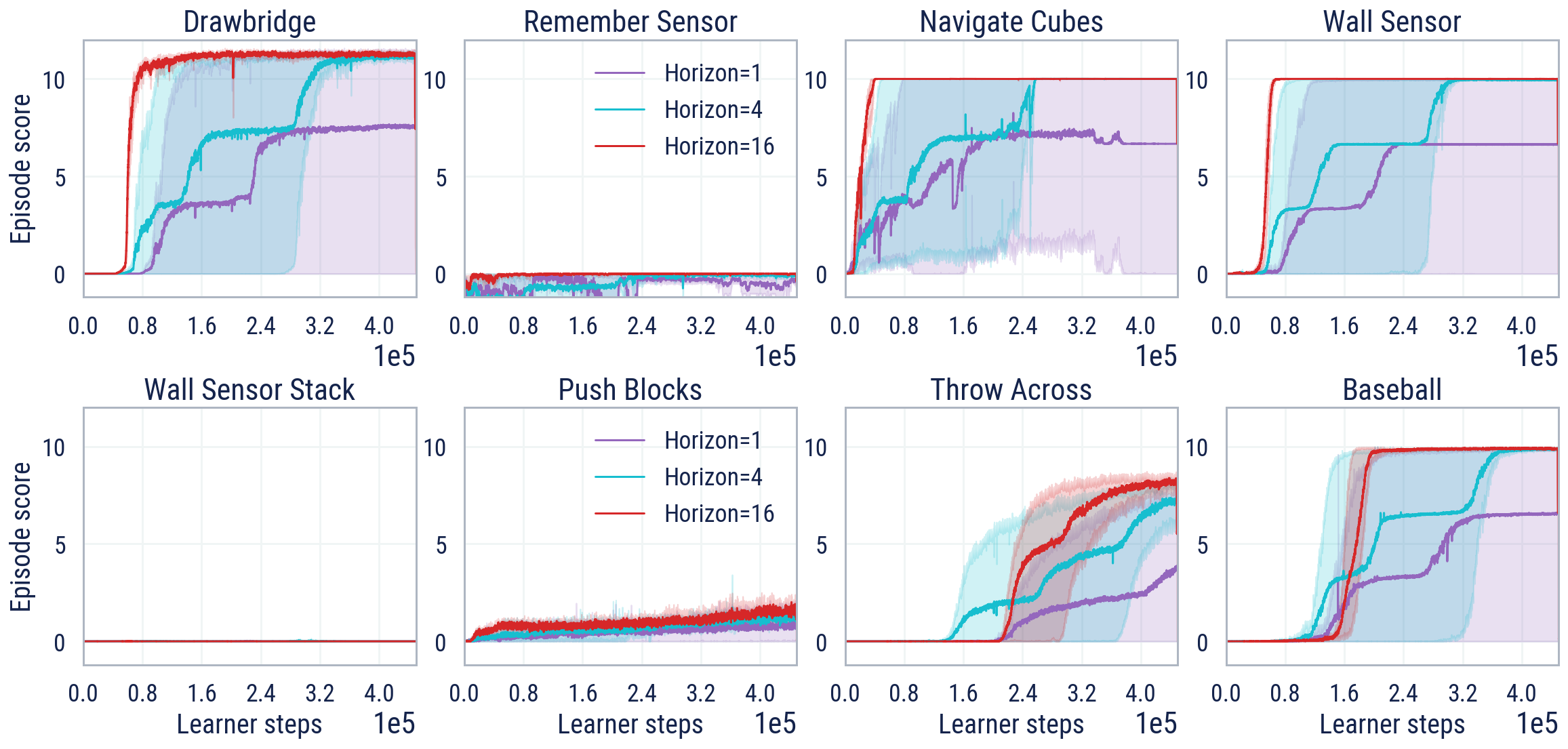}
\caption{Learning curves in terms of agent score for \BYOLE using varying lengths of open-loop horizon $K$ for multistep prediction, for each task in the \DMH suite, averaged over 3 seeds.}
\label{fig:dmh_byole_horizon_ablation}
\end{figure}

\begin{figure}[tb]
\includegraphics[width=1.\textwidth]{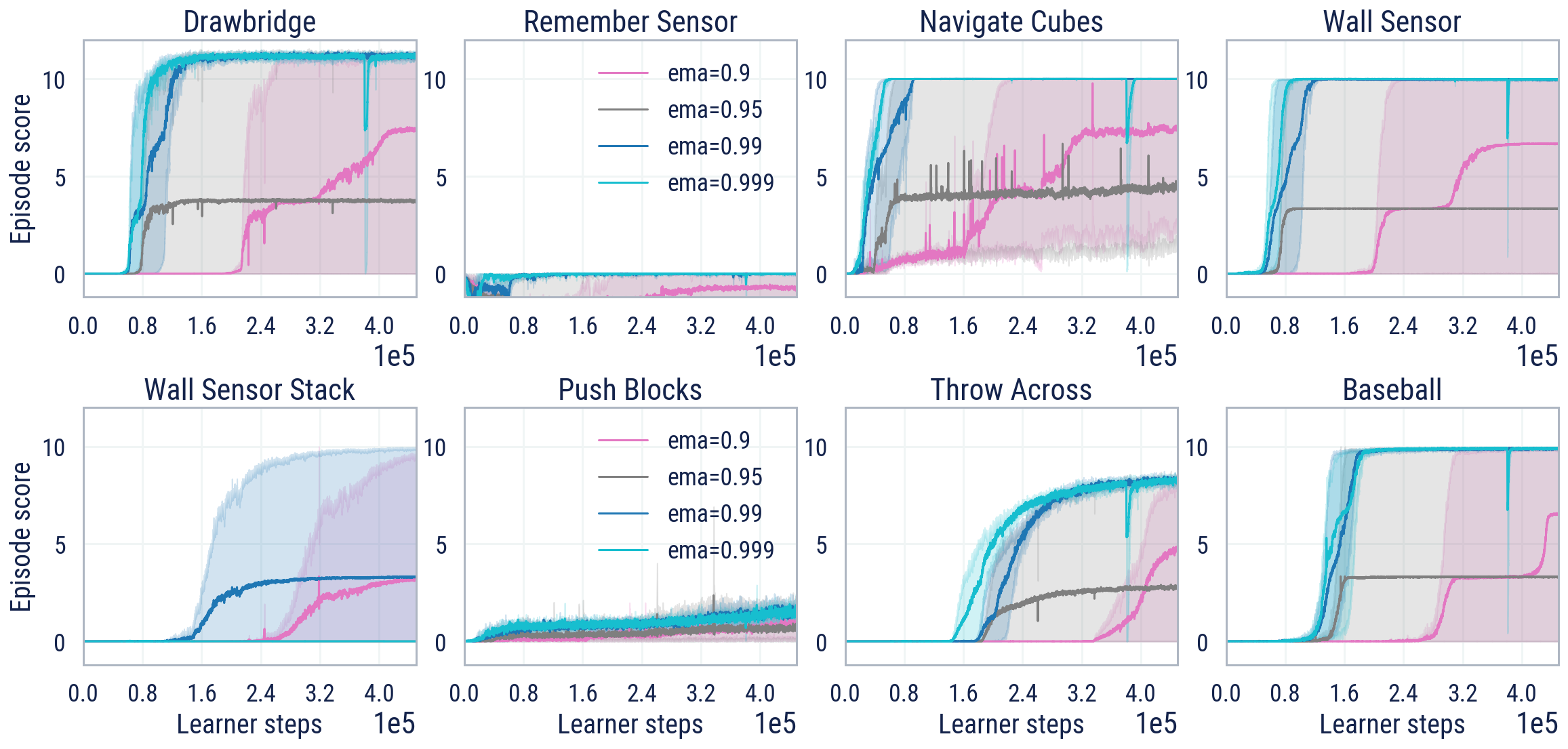}
\caption{Learning curves in terms of agent score for \BYOLE using varying values of EMA target parameter $\alpha$, for each task in the \DMH suite, averaged over 3 seeds.}
\label{fig:dmh_byole_ema_ablation}
\end{figure}

We also conduct further finer-grained ablations of \BYOLE parameters $K$ and $\alpha$. Recall that the main experiments used $K=10$ and $\alpha=0.99$.
Figure~\ref{fig:dmh_byole_horizon_ablation} shows that increasing the open-loop horizon from $K=1$ to $16$ smoothly improves performance, showing the benefits of multi-step prediction of the future and the robustness of \BYOLE to this parameter.
Figure~\ref{fig:dmh_byole_ema_ablation} shows the effect of varying the EMA target parameter $\alpha$. We see that values \BYOLE is fairly robust to this hyperparameter, with the highest performance obtained by our base setting of $\alpha=0.99$.

\begin{figure}[tb]
\includegraphics[width=1.\textwidth]{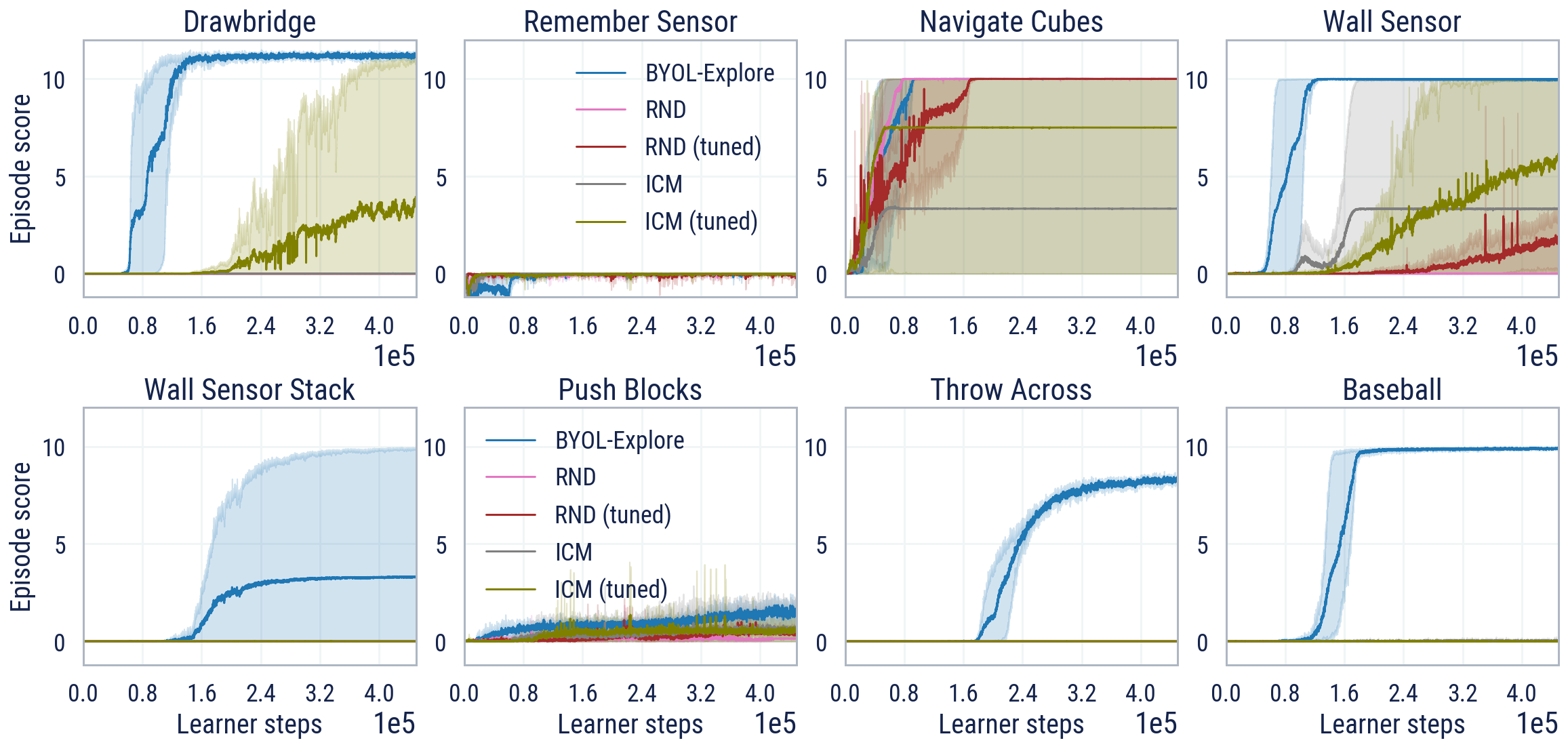}
\caption{Learning curves in terms of agent score for \BYOLE compared against \ICM and \RND both using the same architecture/RL parameters and against baselines tuned separately, for each task in the \DMH suite, averaged over 3 seeds.}
\label{fig:dmh_tuned_baselines}
\end{figure}

Next, we examine different improvements to our main baselines of \RND and \ICM. Note that for ease of comparison, the results reported for the baselines are using the same architectures and VMPO parameters as the \BYOLE settings. Nevertheless, for the sake of steelmanning our baselines, we separately tune both \ICM and \RND to maximize performance and report a comparison in Figure~\ref{fig:dmh_tuned_baselines}. We see that while performance can be improved slightly (at the expense of the tuned \ICM and \RND requiring more parameters), overall \BYOLE still outperforms across the board.

\begin{figure}[tb]
\includegraphics[width=1.\textwidth]{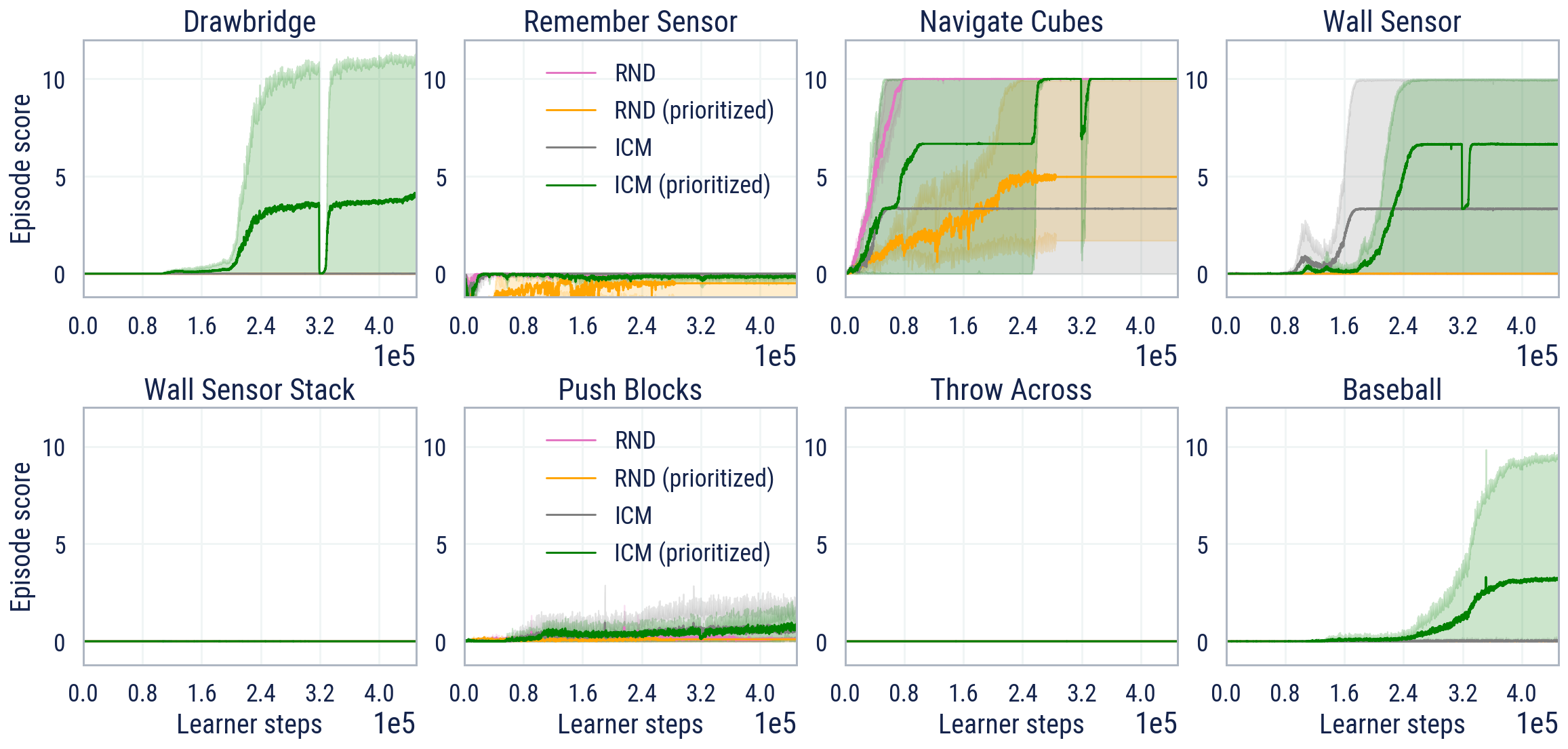}
\caption{Learning curves in terms of agent score for \ICM and \RND both with and without intrinsic reward prioritization via clipping, for each task in the \DMH suite, averaged over 3 seeds.}
\label{fig:dmh_prioritized_baselines}
\end{figure}

Aside from ablating the architectures and RL parameters, we also investigate whether our intrinsic reward prioritization mechanism via clipping can also improve the baselines of \ICM and \RND. Figure~\ref{fig:dmh_prioritized_baselines} shows that, again, while performance can be slightly improved, overall \BYOLE still outperforms baseline. Thus, multistep prediction of bootstrapped latents provides a qualitatively better intrinsic reward, aside from our contribution of a new reward normalization scheme.

\begin{figure}[tb]
\includegraphics[width=1.\textwidth]{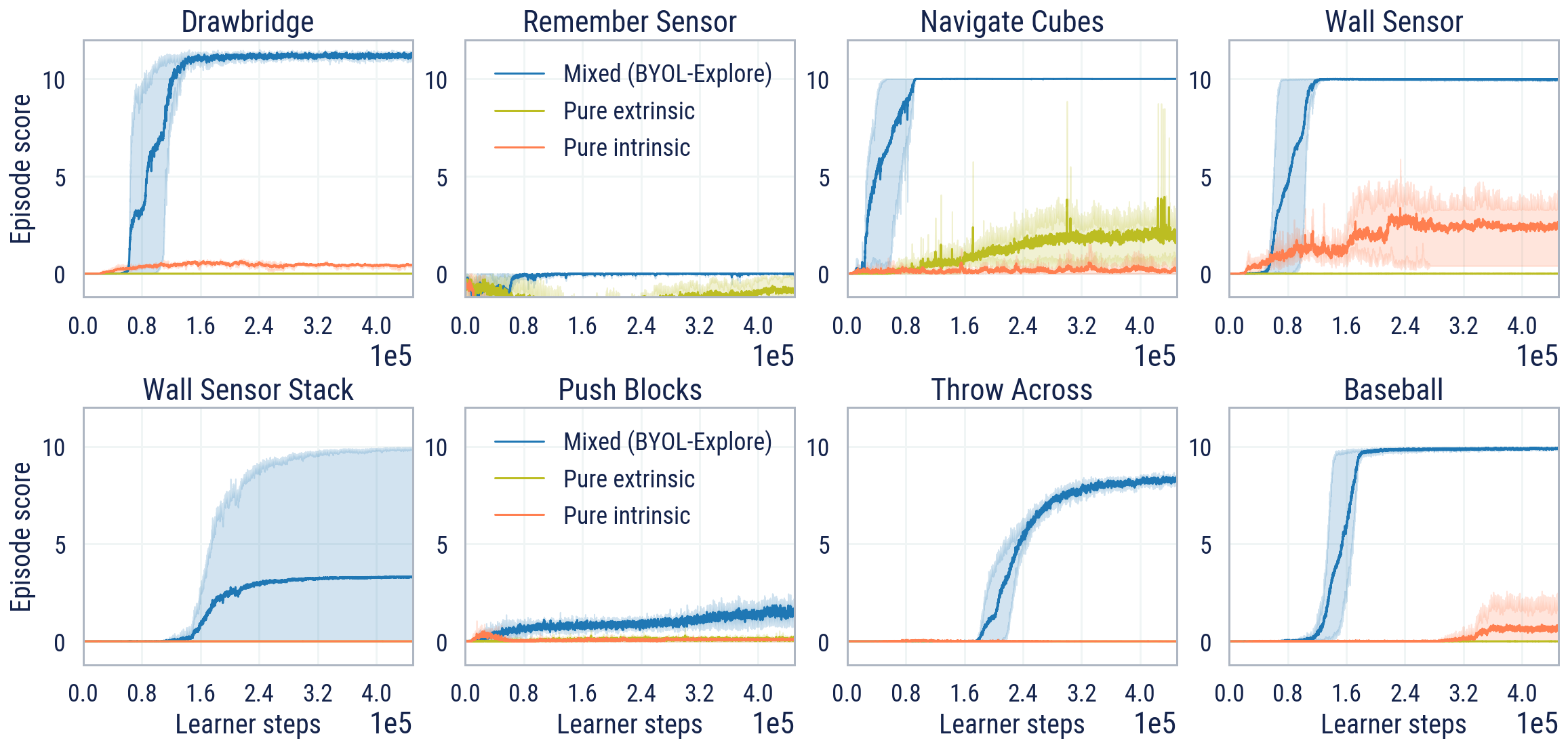}
\caption{Learning curves in terms of agent score for for \BYOLE compared against both pure RL and pure-exploration settings, for each task in the \DMH suite, averaged over 3 seeds.}
\label{fig:dmh_pure_intrinsic}
\end{figure}

Similarly to \texttt{Atari}, in Figure~\ref{fig:dmh_pure_intrinsic}, we also compare our mixed-intrinsic-extrinsic reward optimizing agent against agents optimizing purely intrinsic or extrinsic rewards. In the \DMH domains, due to the much sparser reward than \texttt{Atari}, pure exploration in most tasks does not perform very well. However, remarkably it is still capable of achieving some positive signal on a handful of tasks, particularly \texttt{Wall Sensor}. Thus, purely maximizing intrinsic reward on \DMH still results in meaningful exploration that leads the agent to display interesting behavior.

\begin{figure}[tb]
\includegraphics[width=1.\textwidth]{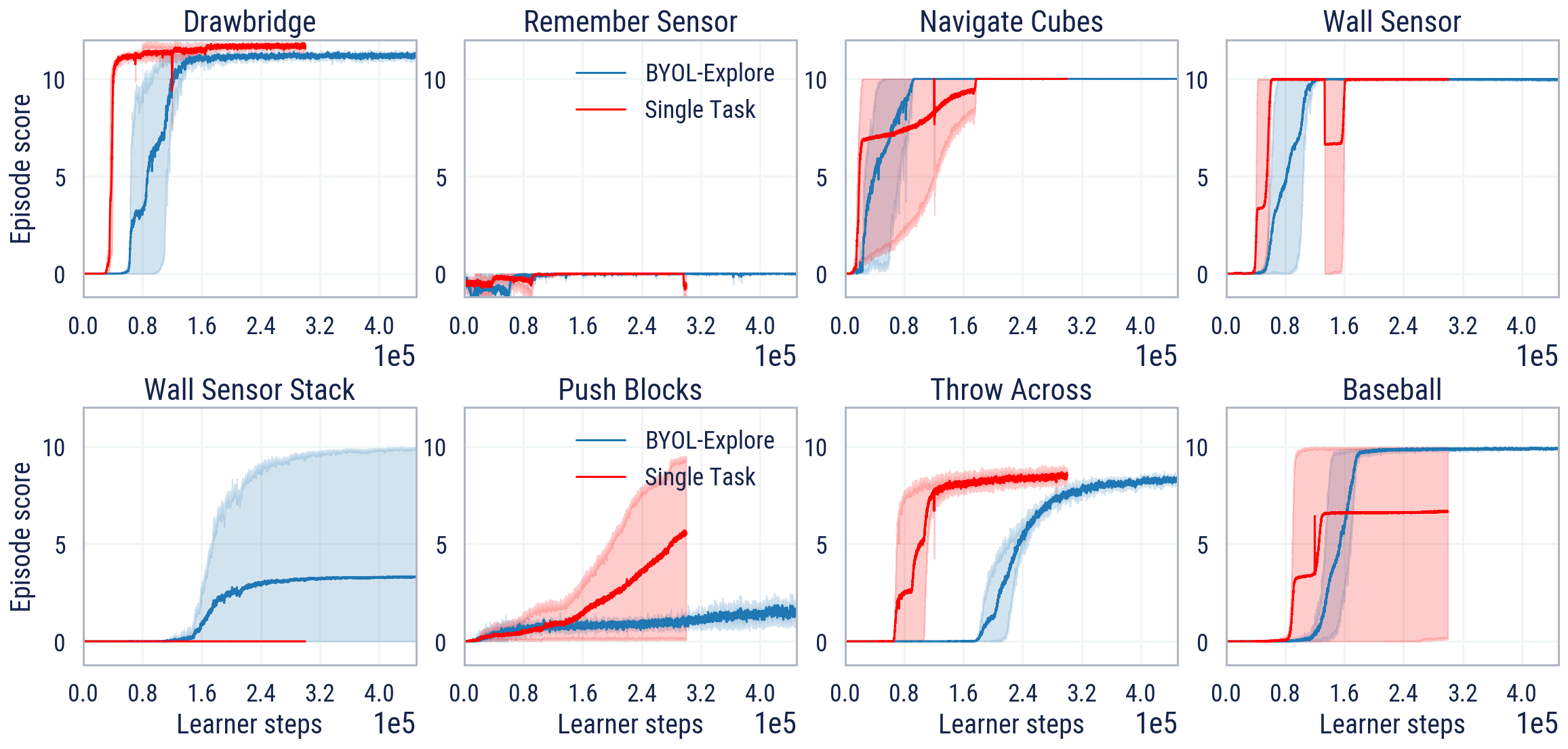}
\caption{Learning curves in terms of agent score for for \BYOLE in the multi-task and single-task setting, for each task in the \DMH suite, averaged over 3 seeds.}
\label{fig:dmh_single_task}
\end{figure}

Finally, we note that we consider the considerably harder multi-task setting compared to the single-task setting chosen in prior work~\cite{gulcehre2019making}. To ensure that the small amount of similarity between tasks does not induce any useful positive transfer and thus make the tasks \textit{easier} when trained on multiple levels, for the sake of completeness we also report in Figure~\ref{fig:dmh_single_task} the performance on single-task setting. As expected, performance is actually slightly higher in the single-task setting, thus showing that our strong multi-task results are not the result of any simplifying changes. Note also that the multi-task setting not only has to learn using a single set of parameters, it also receives $\frac{1}{8}$ of the training data per task in the \DMH suite since we compare based on \textit{total} experience across all tasks being equal.

Note that due to computational constraints we train the single-task version for $320000$ learner steps, but this is sufficient to demonstrate that single-task is not a harder setting than multi-task.

\end{document}